\documentclass{article}

\usepackage{PRIMEarxiv}

\usepackage[utf8]{inputenc} 
\usepackage[T1]{fontenc}    
\usepackage{hyperref}       
\usepackage{url}            
\usepackage{booktabs}       
\usepackage{amsfonts}       
\usepackage{nicefrac}       
\usepackage{microtype}      
\usepackage{fancyhdr}       
\usepackage{graphicx}       
\graphicspath{{media/}}     
\usepackage{multicol}  
\usepackage{multirow}
\usepackage{soul}
\usepackage{graphicx}
\usepackage{textcomp}
\usepackage{xcolor}

\usepackage{tabularx}       
\usepackage{subfig}         
\usepackage{enumitem}       
\usepackage{amsmath}        

\usepackage{algorithm}
\usepackage{algorithmic}

\newcommand{\q}[1]{``#1''}

\newcolumntype{Y}{>{\centering\arraybackslash}X}

\title{POPNASv3: a Pareto-Optimal Neural Architecture Search Solution for Image and Time Series Classification
}

\author{
  Andrea Falanti, Eugenio Lomurno, Danilo Ardagna, Matteo Matteucci  \\
  Politecnico di Milano \\
  Milan, Italy\\
  \texttt{\{andrea.falanti, eugenio.lomurno, danilo.ardagna, matteo matteucci\}@polimi.it} \\
}

\begin{document}
\maketitle

\begin{abstract}
The automated machine learning~(AutoML) field has become increasingly relevant in recent years.
These algorithms can develop models without the need for expert knowledge, facilitating the application of machine learning techniques in the industry.
Neural Architecture Search~(NAS) exploits deep learning techniques to autonomously produce neural network architectures whose results rival the state-of-the-art models hand-crafted by AI experts.
However, this approach requires significant computational resources and hardware investments, making it less appealing for real-usage applications.
This article presents the third version of Pareto-Optimal Progressive Neural Architecture Search~(POPNASv3), a new sequential model-based optimization NAS algorithm targeting different hardware environments and multiple classification tasks.
Our method is able to find competitive architectures within large search spaces, while keeping a flexible structure and data processing pipeline to adapt to different tasks.
The algorithm employs Pareto optimality to reduce the number of architectures sampled during the search, drastically improving the time efficiency without loss in accuracy.
The experiments performed on images and time series classification datasets provide evidence that POPNASv3 can explore a large set of assorted operators and converge to optimal architectures suited for the type of data provided under different scenarios.
\end{abstract}



\section{Introduction}
Machine learning~(ML) is currently the most researched topic in the artificial intelligence~(AI) field, since its application can solve and automatize tasks difficult to address with standard software programming.
Over the past decade, deep learning~\cite{goodfellow2016deep, Alzubaidi2021} has gradually become the leading field in AI applications.
Indeed, neural networks can take advantage of the huge amount of data gathered by modern information systems and the advancement in parallel computation technology to effectively learn multiple tasks, like image classification, speech recognition, time series forecasting, and many others~\cite{Dargan2020}.
These complex models have progressively become the state-of-the-art solution in different domains, starting with image classification of which famous architecture examples are AlexNet~\cite{AlexNet}, VGG~\cite{VGG} and ResNet~\cite{ResNet}.

Machine learning techniques make it possible to automatically learn from data, but AI experts must design effective models and training procedures to fit the data properly and generalize the predictions for real-world scenarios.
In the deep learning field, this stage is referred to as \emph{architecture engineering} and it addresses the choice of layers used in the neural network model, the number of layers to stack, their hyperparameters (i.e. filters, kernel sizes), and other training details, like the best loss function for the task and the regularization techniques to use.
This process is tedious and time-consuming, because the model designer must iteratively perform multiple experiments and adapt the architecture based on the obtained results.
Furthermore, the results of neural networks are difficult to interpret and explain, making it unintuitive how to modify the models to achieve performance improvements. 

Automated machine learning~(AutoML) represents the solution to automatize the model design step, without the need for human expertise.
It has recently gained large interest in the research world~\cite{DBLP:journals/corr/abs-1904-12054}, since it could be one of the next breakthroughs in the AI field, facilitating mass adoption of ML in the industry.
Given the power and adaptability of neural networks, this approach has also been studied in deep learning and it goes under the name of \emph{neural architecture search}~(NAS).
The first successful application of NAS has been published by Zoph \emph{et al.}~\cite{zoph2017neural}, addressing the search as a reinforcement learning~(RL) problem, where a controller is able to output network structures and learn from their results.
Even if the performances of the final networks were in-line with state-of-the-art handcrafted architectures, the search required multiple GPUs and more than 30000 cumulative GPU hours.

To make NAS successful in real-world scenarios, the search must be performed in a reasonable amount of time, without the need for large hardware investments.
Still, the search space and methodology should be flexible enough to address different tasks and datasets, in order to generalize performance to problems not seen during algorithm design.
Successive NAS works focused more on search efficiency and integrated multi-objective optimization to handle potential deployment constraints.
A popular technique is to exploit One-shot models~\cite{OneShotUniform, UnderstandingOneShot}, where a single huge neural network, named supernet, represents the entire search space expressed by the algorithm.
The main advantage of using a supernet is that its number of layers scales linearly with the search space dimension, while the number of architectures scales exponentially with the number of operators and inputs to combine.
Despite the efficiency benefits, supernets are challenging to optimize since paths tend to co-adapt their weights and they are also more sensitive to weight initialization and hyperparameter settings.
Moreover, the search space is limited by the memory capacity of the training device, since the whole supernet must be stored in memory during the search procedure.

In this work, we propose a sequential model-based optimization~\cite{smbo} method called POPNASv3, whose workflow can be adapted to different tasks and data sources.
The algorithm can be applied to image classification and time series classification with minimal changes in its configuration, supporting vast search spaces without sacrificing either time efficiency or performance in classification metrics.
Results indicate that the top architectures found by our method compete with other state-of-the-art methods, having slightly lower accuracy but great time efficiency while exploring larger search spaces than other NAS algorithms, on average.
The search space can be potentially unbound with our progressive approach and POPNASv3 does not require any pretraining, making it a good candidate for fast prototyping and for studying the most promising combinations of operators out of a vast operator set. 

The key contributions of this paper are:
\begin{itemize}[noitemsep]
    \item the extension of POPNAS workflow to time series classification tasks, adding operators relevant to this domain like RNN units and dilated convolutions and adapting the already supported operators for the new input shape.
    \item the addition of two post-search procedures, namely the \emph{model selection} and \emph{final training} steps.
    The model selection is performed on the top-5 architectures found during the search, training them more extensively and automatically tuning their macro-configuration. The final training is performed on the best configuration found during model selection, training the model to convergence.
    \item the adaptation of the training procedure to support different training devices, making now possible to train each network also on multiple GPUs, through synchronous variable replicas, and on TPUs.
    We demonstrate that the time predictor is able to learn how the training time of the sampled networks varies on each device, making the algorithm robust to the different training strategies.
    \item the possibility of using residual connections on cell outputs and employing fine-grained adjustments to alter the tensor shapes of each operator, making it possible to freely mix heterogeneous operators inside the architectures. 
\end{itemize}

\section{Related Works}
Neural architecture search~(NAS) is a fast-evolving research field in the deep learning community, which combines a multitude of different techniques studied in machine learning.
In this section, we provide a general description of NAS and summarize the intuitions behind the most endorsed works of the literature, on which the majority of subsequent algorithms are based.

\subsection{Neural Architecture Search}
The goal of neural architecture search is to autonomously design complex neural network architectures in a single end-to-end process, which does not require human intervention.
NAS is a young but popular field of research among the AI community, which still lacks proper standardization, given also the large number of techniques involved in these works.
A tentative classification of NAS algorithms is provided by~\cite{elsken2019neural}, which characterizes the methods according to three main traits:
\begin{itemize}[noitemsep]
    \item the \emph{search space}, which defines the architectures set expressed by the algorithm
    \item the \emph{search strategy}, which defines how the algorithm samples the search space to find the optimal architectures for the considered task
    \item the \emph{performance estimation strategy}, which defines how to evaluate the architectures' quality with a minimal time investment
\end{itemize}

Early NAS works~\cite{zoph2017neural, real2019regularized, zoph2018learning}, while successful from an architecture quality standpoint, required enormous amounts of GPU time to find optimal architectures, making the application of NAS in real-world contexts prohibitively expensive.
Several solutions and techniques have been considered for implementing each NAS trait, aiming to reduce the search time while preserving the final accuracy achieved by the architectures and potentially improving it further.

Regarding the search space, each method typically defines a set of operators and how to interconnect them, plus the number of layers or the maximum depth of the architectures generated.
The number of architectures expressed by a defined search space is exponential to its number of operators, layers and possible inputs per layer, making the search space cardinality explode even for small-sized architectures.
NASNet~\cite{zoph2018learning} proposed a cell-based approach, where the search focuses on normal cells and reduction cells, which are basic units aggregating multiple operations.
This idea takes inspiration from famous models, like ResNet~\cite{ResNet} and InceptionNet~\cite{Inception}, where a module is repeated multiple times to build the final architecture.
Similarly, the cells found by NASNet are repeated multiple times inside the final neural network architectures, which is effective for restricting the cardinality to feasible sizes compared to searching freely the entire architecture structure.
Thus, most algorithms focus their search effort on the \emph{micro-architecture}, founding good cells for the task, while the \emph{macro-architecture}, i.e. how and how many these cells are stacked together, is often tuned after the search.

Concerning the search strategy and performance estimation strategy, there are instead very heterogeneous proposals in the literature.
PNAS~\cite{liu2018progressive} proposes a sequential model-based optimization~(SMBO)~\cite{smbo} technique, starting with shallower models which are progressively expanded with new layers, based on the indication of a surrogate model, referred to as \emph{predictor}.
The predictor gives an estimate of the accuracy reachable by each candidate architecture and is fitted on the training results of the networks previously sampled at runtime, which makes it resilient to search space and hardware modifications.
Works like AmoebaNet~\cite{real2019regularized} and NSGANet~\cite{NSGANet} employ evolutionary algorithms to perform the architecture search, while the weights are optimized with standard gradient descent techniques.
In evolutionary algorithms, the search space represents the \emph{phenotype}, while the architectures searched by the algorithm are mapped to \emph{genotypes} through an encoding.
These works retain a population of architectures in each iteration, whose genotypes are modified with mutation and crossover operations to generate an offspring.
Mutations alter the encoding by randomly flipping bits, which in NAS encodings results often in adding/removing a layer or adding/removing a connection between two layers.
Crossover operations instead mix two encodings to build new architectures which share characteristics from their parents.
ENAS~\cite{ENAS} increases the search efficiency by combining a reinforcement learning search strategy with weight sharing, since the weights learned by previously trained architectures can be re-used in similar ones to speed up the training phase.
Multiple works started also to consider NAS as a multi-objective optimization problem~\cite{elsken2018efficient, hsu2018monas, POPNAS, POPNASv2}, addressing potential deployment constraints and increasing the search efficiency without trading-off accuracy.

DARTS~\cite{liu2019darts} improves the search efficiency by applying a relaxation on the search space to make it continuous.
In this way, the entire search space can be represented as a single huge network, i.e. the \emph{supernet}, where each edge between a pair of layers $(i, j)$ is parametrized by a continuous variable $ \alpha_{ij} $.
This change makes it possible to optimize both the weights and the model structure using gradient descent, alternating two specialized training steps.
The final architecture is a subgraph of the supernet, selected by choosing the paths with the maximum parameters $ \alpha $ at each branching.
DARTS efficiency results have contributed to the popularity of a new line of methods based on One-shot models~\cite{OneShotUniform, UnderstandingOneShot}, in which the supernet is defined and pretrained prior to performing the actual neural architecture search.
During the search, any architecture sampled by the algorithm can be seen as a composition of paths of the supernet, whose weights are taken from the ones learned by the supernet and just fine-tuned, to get an extremely fast evaluation of the network accuracy.
Still, training supernets is challenging since they are more sensitive to hyperparameters changes and weight co-adaptation, requiring peculiar training techniques to not favor particular architectures during the final estimation phase.
The memory capacity of the training device limits the search space cardinality as the whole network must reside on the device, but techniques like elastic kernels and progressive shrinking proposed by Once-for-All~\cite{cai2020OFA} can mitigate this issue.
Furthermore, one-shot models still require a considerable amount of GPU hours for pre-training, which is amortized only if they can be reused multiple times during different NAS problems.

\subsection{Time Series Classification}
The majority of NAS algorithms are designed to solve image classification problems and only rarely extend their search procedures to different domains.
Time series classification~(TSC) tasks introduce new challenges due to the plethora of data sources used in this field and the different data correlations exploitable with this kind of data.

Time series classification consists in predicting the correct label for each input series among multiple label choices.
TSC datasets can be univariate, when they have a single feature per timestep, or multivariate, when they have multiple features per timestep.
Each feature represents a different time series, from which the model can learn potential correlations for predicting the label.
The data sources of time series samples are heterogeneous: for example, time series can represent audio, positional data, and sensors from medical devices.
These types of signals are sampled at different frequencies and can exhibit behaviors in either the long or short term.
Furthermore, these datasets tendentially suffer from more data scarcity due to privacy issues, compared to the ones used for image classification purposes.
These characteristics make it much harder to find a single optimal model which can fit well data from any domain.

In the literature, TSC tasks have been addressed via different ways to represent temporal patterns, with the goal of extracting useful information.
The most common approaches are distance-based, dictionary-based, interval-based and shapelet-based techniques.
State-of-the-art methods aggregate multiple data representations, making them robust to different application tasks. Most of these works have been tested on the UCR~\cite{UCRArchive2018} and UEA~\cite{UEAArchive2018} archives, which are curated groups of datasets commonly used as benchmarks for evaluating the performance on multiple TSC domains.
InceptionTime~\cite{InceptionTime} is one of the most popular deep learning methods for TSC, which relies on learning an ensemble of convolutional neural networks.
Each model shares the same structure, applying in parallel convolution with different kernel sizes to extract features from both short and long time windows.

ROCKET~\cite{ROCKET} is another approach based on convolutional operators, but interestingly, it does not train the weights like in deep learning methods, simply generating them randomly.
ROCKET demonstrates that it is possible to extract meaningful features from completely different datasets by just using an order of 10000 randomly-initialized convolutions with different properties.
To produce the final output, a linear classifier is trained on the maximum value and the proportion of positive values extracted by each convolution, making ROCKET significantly faster to train than other state-of-the-art methods, while obtaining similar performances.
The authors noticed that dilated convolutions greatly contribute to the effectiveness of the algorithm, which is a valuable insight for designing convolutional neural networks for these tasks.

Considering only the average accuracy, HIVE-COTEv2~\cite{HIVE-COTEv2} is currently the best-performing method on the UCR and UEA archives.
HIVE-COTEv2 is based on 4 different classifiers, which are themself ensembles of multiple techniques and models.
The final output is built through a control unit which re-weights the probabilities produced by each classifier, based on an estimate of their quality.
Even if HIVE-COTEv2 performs better than the previously cited methods, it is computationally slower since it trains more models and it has no support for GPU parallelization.
In the NAS literature, the only work focusing on TSC is NAS-T~\cite{NAS-T}, which uses an evolutionary algorithm to fine-tune a pre-trained ResNet~\cite{ResNet-for-TSC} model with additional layers, taken from the defined search space.
NAS-T stops after 5000 network evaluations, or 48 hours of GPU time, on each dataset of the UCR archive.
The results obtained in the experiments indicate that NAS-T is a marginal improvement over the first version of HIVE-COTE.

\section{Method}
Reaching results similar to the state-of-the-art architectures in generic tasks, with reasonable hardware and GPU time investment, is crucial for applying NAS in the industry.
POPNASv3 goal is to provide fast neural architecture prototyping, optimizing the computation time without any significant accuracy drop compared to more costly NAS methods.
To do so, we extended the POPNASv2~\cite{POPNASv2} method, which was inspired by the original POPNAS~\cite{POPNAS} and PNAS~\cite{liu2018progressive} ones.
The algorithm proposed is characterized by a sequential model-based optimization~(SMBO) strategy, which is enriched by the presence of accuracy and time predictors to estimate respectively the accuracy and training time of candidate architectures.
Thus, our work addresses NAS as a time-accuracy optimization problem, selecting only the architectures which belong to the Pareto front computed on the estimates made by the predictors, since these models achieve the best tradeoff between the two metrics.
Pareto optimization can reduce the number of networks to sample in each step, while preserving a set of time-efficient diversified architectures.
Training structurally different architectures helps the predictors in finding the key aspects which contribute to the gaps in training time and accuracy, progressively improving the Pareto front accuracy at each iteration.
In this section, we provide a comprehensive explanation of how POPNASv3 implements the three main NAS traits, namely the search space, search strategy and performance estimation strategy, in order to achieve the intended goal.

\subsection{Search Space}

\begin{figure}[t]
	\centering
	\subfloat[block structure]{
	    \label{FIG:block}
	    \includegraphics[width=0.21\textwidth]{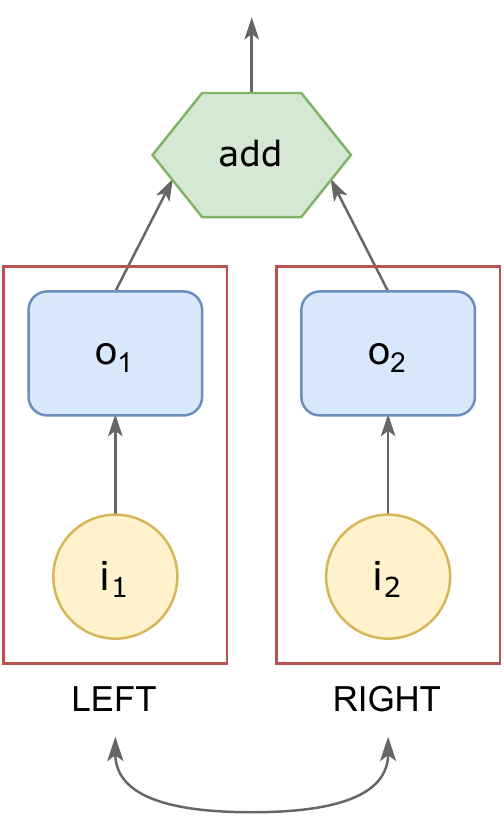}
	}
	\hfill
	\subfloat[cell structure example]{
	    \label{FIG:cell}
	    \includegraphics[width=0.25\textwidth]{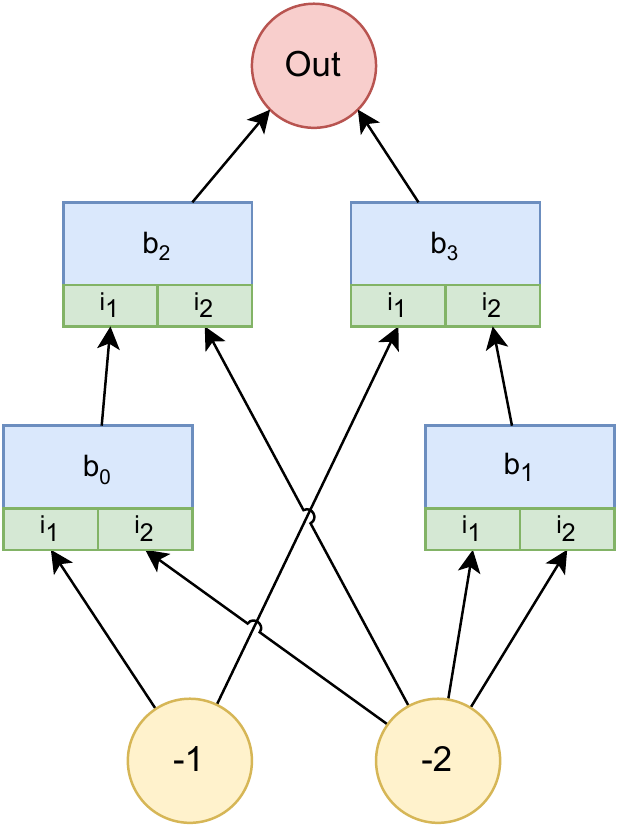}
	}
	\hfill
	\subfloat[macro-architecture overview]{
	    \label{FIG:macro-arch}
	    \includegraphics[width=0.35\textwidth]{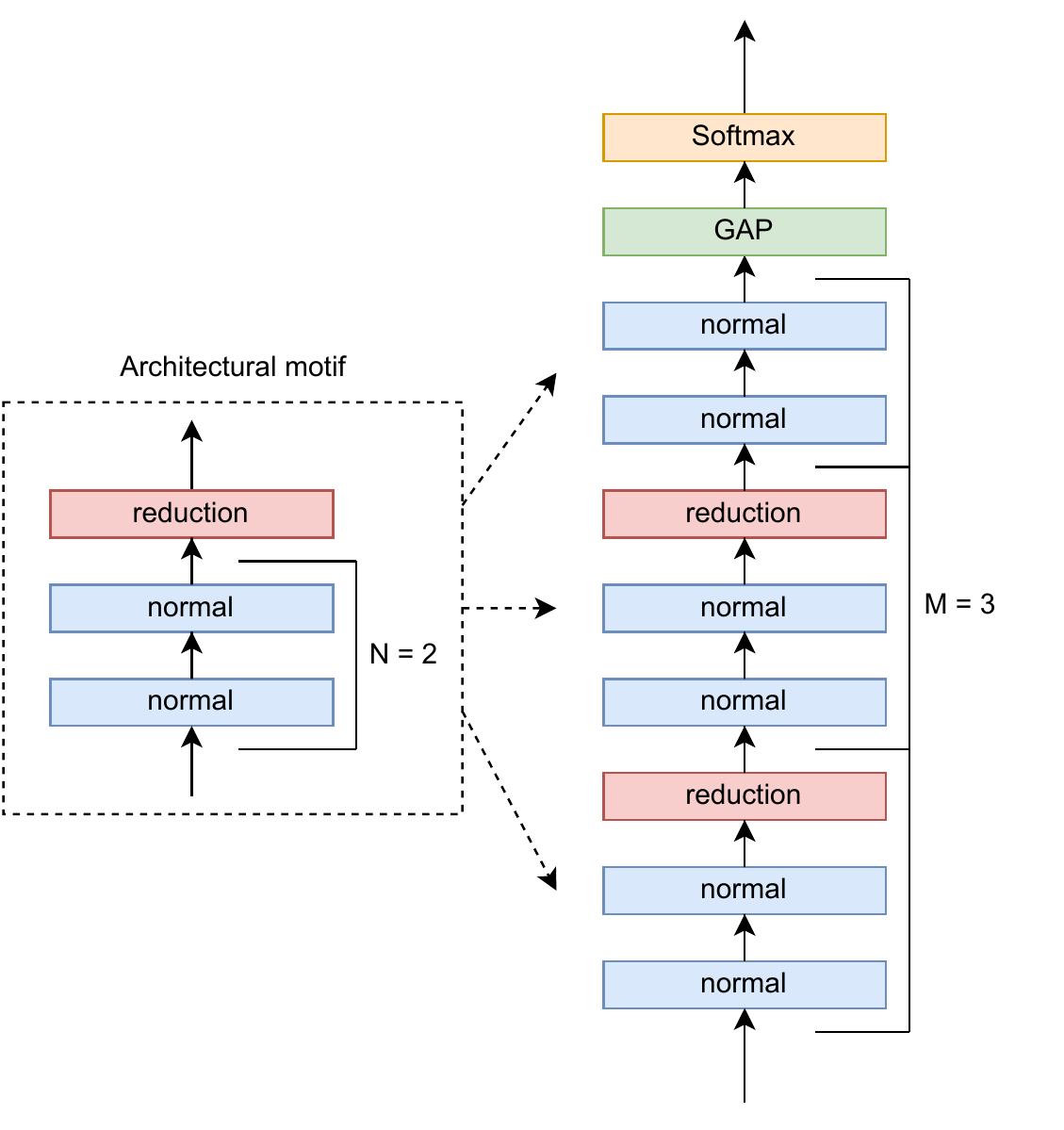}
	} 
	\caption{An overview of all the structural units defined in POPNASv3: block, cell, and architectural motif. (a) A block is the combination of two operators into a single output. Since POPNASv3 always use add as the join operator, which is commutative, LEFT-RIGHT and RIGHT-LEFT blocks are isomorphic and therefore only one of them is generated. (b) An example of cell specification, which is a DAG composed of multiple interconnected blocks, having a single output. (c) The standard macro architecture used by POPNASv3 during the search process. An architectural motif is composed of \textit{N} normal cells followed by a reduction cell. \textit{M} motifs are stacked to build the final architecture, followed by GAP and Softmax as standard in classification networks.}
	\label{FIG:micro-macro-arch}
\end{figure}

The set of architectures included in the search space is defined by the chosen operator set $ \mathcal{O} $ and the set of hidden states $ \mathcal{I} $ which can be used as input in the cell.
Our work uses a cell-based approach, composing each cell with multiple micro-units called blocks, similar to what is done in \cite{liu2018progressive,zoph2018learning}.
A block~(Figure~\ref{FIG:block}) is specified with a tuple $ (in_{1}, op_{1}, in_{2}, op_{2}) $, with $ in_{1}, in_{2} \in \mathcal{I} $ and $ op_{1}, op_{2} \in \mathcal{O}$, where $ in_{1} $ is processed by $ op_{1} $ and $ in_{2} $ is processed by $ op_{2} $.
The new tensors are merged together with a join operator, which in our case is fixed to the add operation to keep constant the number of output channels.
A cell~(Figure~\ref{FIG:cell}) is simply the composition of multiple blocks, nested as a directed acyclic graph~(DAG).
Following PNAS original intuitions, POPNASv3 defines two types of cells: \emph{normal} cells, which preserve the dimensionality of the input, and \emph{reduction} cells, which halve the spatial resolution and double the filters.
Normal and reduction cells share the same structure to maximize the search efficiency: the only difference between them is that all the operators of a reduction cell using a lookback as input perform a stride of factor 2, to reduce the spatial dimensionality.
To make the shape adaptation possible, operators that can not alter the spatial dimensionality or the number of filters are accompanied by respectively max pooling and pointwise convolution layers, but only when the shape is effectively adjusted.

POPNASv3 focuses mainly on convolutional and pooling operators.
The operators are defined in the apposite section of the configuration file and are recognized through regexes, making it possible to parametrize the kernel sizes and dilation rates for different tasks.
Taking as an example the 2D convolution operator, it is specified as the regex \verb|"\d+x\d+(:\d+dr)? conv"|, where the first two numbers define the kernel size and the optional third one the dilation rate, set to 1 by default.
For time series, it is also possible to define some popular RNN operators, namely LSTM~\cite{hochreiter1997long} and GRU~\cite{GRU}.
All supported operators are:
\begin{itemize}[noitemsep]
    \item identity
    \item @x@(:@dr)? conv
    \item @x@(:@dr)? dconv (Depthwise-separable convolution)
    \item @x@-@x@ conv (Spatial-separable convolution)
    \item @x@ tconv (Transpose convolution)
    \item @x@ maxpool
    \item @x@ avgpool
    \item LSTM (time series only)
    \item GRU (time series only)
\end{itemize}
where \emph{@} can be replaced by any number.
Note that the convolutional and pooling operators can be used in time series classification by just adapting their kernel size to be monodimensional (e.g @ conv, @ maxpool).
All convolutional operators are implemented as a sequence of convolution, batch normalization~\cite{BatchNorm} and Swish~\cite{ramachandran2017searching} activation function.
Most NAS methods use ReLU as activation function since it is immediate to derive, but, in our training procedure, it rarely caused learning difficulties in shallower architectures due to the dying neurons problem.
From our preliminary studies, Swish seems more resilient to the learning rate and weights initialization, helping in getting more accurate estimations of the quality of all the networks included in the search space.
However, we noticed that using Swish results in an increase in network training time of more than 10\% compared with ReLU, which is a significant trade-off to pay in order to maximize the search stability for general usage.

The input set can be altered by defining the maximum allowed \emph{lookback} input, i.e. the maximum distance in cells from which the current cell can take hidden states (outputs of previous cells).
For example, the default lookback distance in the POPNASv3 configuration is 2: this means that a cell can use the output of the previous cell (-1 lookback) and the second to last cell (-2 lookback) as input for its blocks.
In addition to lookback inputs, each block in a cell can use any previous block in the cell specification as input, making it possible to nest blocks in complex multilevel DAGs.
Indeed, each cell can be considered as a DAG using lookbacks as inputs and terminating with a single output node.
All unused block outputs are concatenated into the cell output, followed by a pointwise convolution to keep the number of filters consistent throughout the network.
If enabled in the experiment configuration, the cell output can be made residual by summing the cell output with the nearest lookback input utilized by the cell.

Regarding the macro architecture (Figure~\ref{FIG:macro-arch}), we group the cells in units called \emph{architectural motifs}.
Each architectural motif is composed of $ N $ subsequent normal cells, followed by a single reduction cell.
At search time, the algorithm stacks $ M $ motifs to build each architecture.
POPNASv3 focuses on the micro-architecture search, while the macro-architecture parameters are tuned afterward in the model selection phase.

\subsection{Performance Estimation Strategy} \label{SEC:perf-estimation}%
All architectures sampled by the algorithm are trained for a short amount of epochs, to obtain a quick estimate of the quality reachable during prolonged training.
This technique is often referred to in the NAS literature as \emph{proxy training}.
The training procedure is carried out via the AdamW~\cite{loshchilov2019decoupled} optimizer with a cosine decay restart~\cite{loshchilov2016sgdr} schedule for both the learning rate and weight decay, which provides fast convergence and allows to quickly distinguish the most promising architectures from suboptimal ones.
Two surrogate models, referred to as accuracy predictor and time predictor, guide the algorithm in selecting the most promising architectures in each step.

\subsubsection{Accuracy Predictor} \label{SEC:acc-predictor}%

\begin{figure}[t]
	\centering
	\includegraphics[width=0.75\textwidth]{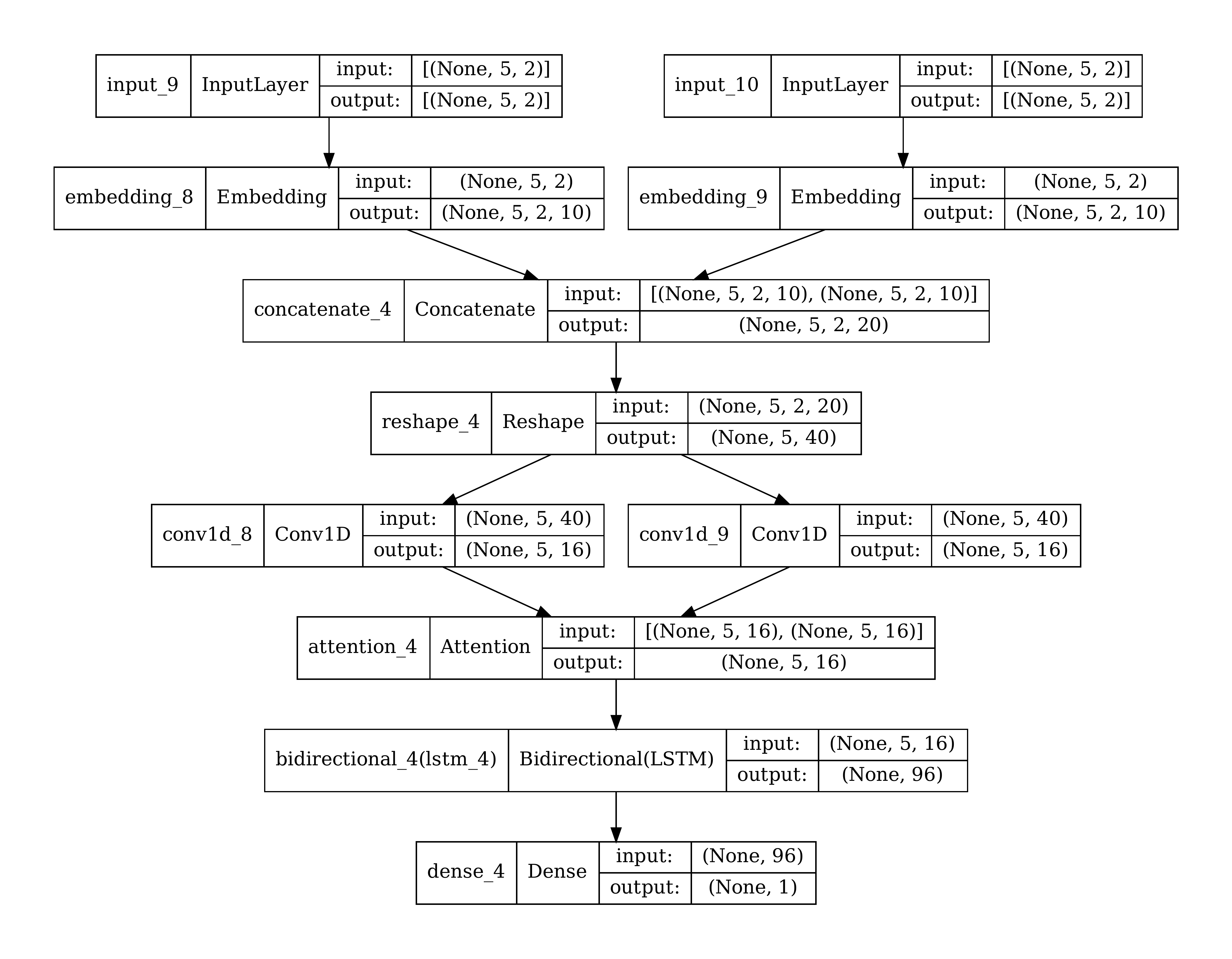}
	\caption{The neural network structure designed for POPNASv3 accuracy predictor.}
	\label{FIG:acc-predictor}
\end{figure}

The accuracy predictor is implemented as an LSTM~\cite{hochreiter1997long} with Self-Attention~\cite{SelfAttention}.
This type of network can directly process the cell specification, i.e. the list of blocks, as features.
The entire structure of the accuracy predictor is illustrated in Figure~\ref{FIG:acc-predictor}.
The predictor expects two inputs: the list of all cell inputs and the list of all cell operators, both encoded as 1-indexed integer categorical values.
These tensors have a shape of \emph{(B, 2)} since each block has two inputs and two operators.
Architectures with less than \emph{B} blocks are padded with zeros.
The input tensors are embedded separately and then reshaped together to extract the block features.
This new hidden state is processed in parallel by two convolutions, outputting the \emph{Q} and \emph{K} matrices of the Self-Attention layer.
Finally, the Attention layer is followed by a bidirectional LSTM and a single sigmoid unit, whose output represents the expected accuracy of the architecture during the proxy training.

\subsubsection{Time Predictor} \label{SEC:time-predictor}%

POPNASv3 time predictor is a CatBoost~\cite{prokhorenkova2019catboost} regressor, a state-of-the-art ML method based on gradient-boosting and decision trees.
In contrast to the accuracy predictor case, we extrapolate a specific set of features from the cell structure, instead of passing directly the cell encoding to the model.
The features selected for the predictor are:
\begin{itemize}[noitemsep]
    \item the number of blocks
	\item the number of cells stacked in the final architecture
	\item the sum of the dynamic reindex value of all operators inside the cell (OP score)
	\item the number of concatenated tensors in cell output (number of unused blocks in DAG cell)
	\item use of multiple lookbacks (boolean)
	\item the cell DAG depth (in blocks)
	\item the number of block dependencies
	\item the percentage of the total OP score involved in the heaviest cell path
	\item the percentage of the total OP score due to operators using lookbacks as input
\end{itemize}

These features are associated with the structural changes happening in the cell DAG and the efficiency of the operators on the training device, making them suitable for any dataset, problem and hardware environment.
In particular, the so-called \emph{dynamic reindex} is a formula we designed to depict the time efficiency of each operator $ o \in \mathcal{O} $ compared to the others.
The dynamic reindex function maps each operator with a value between 0 and 1, which is proportional to the estimated time impact of the considered operator compared to the one estimated to take the most time.
These values are computed at runtime, after the algorithm terminates the training of architectures composed of one block.
Receiving as input the set of training times $ \mathcal{T} = \{t_{o} | \forall o \in \mathcal{O}\} $ of the mono-block cells with symmetric encoding $[(-1, o, -1, o)]$, and the training time $ t_{0} $ of the empty cell, the dynamic reindex formula is expressed as:
\begin{equation*}
	dyn\_index_{o} = \dfrac{t_{o} - t_{0}}{max(T) - t_{0}}.
\end{equation*}
Empty cell $ t_{0} $ is considered as a bias, representing the computational effort required by elements common to all architectures of the search space.
The training time due to data preprocessing, data augmentation, and other training-related events, plus the time investment for training the output layers included in all architectures (GAP and Softmax) is detected by $ t_{0} $ and therefore excluded from the dynamic reindex values.
This expedient allows the time predictor to learn more accurately the computational cost differences between all the operators of the search space.

\begin{figure}[t]
	\centering
	\subfloat[SHAP importance per feature]{
	    \label{FIG:shap-importance}
	    \includegraphics[width=0.45\textwidth]{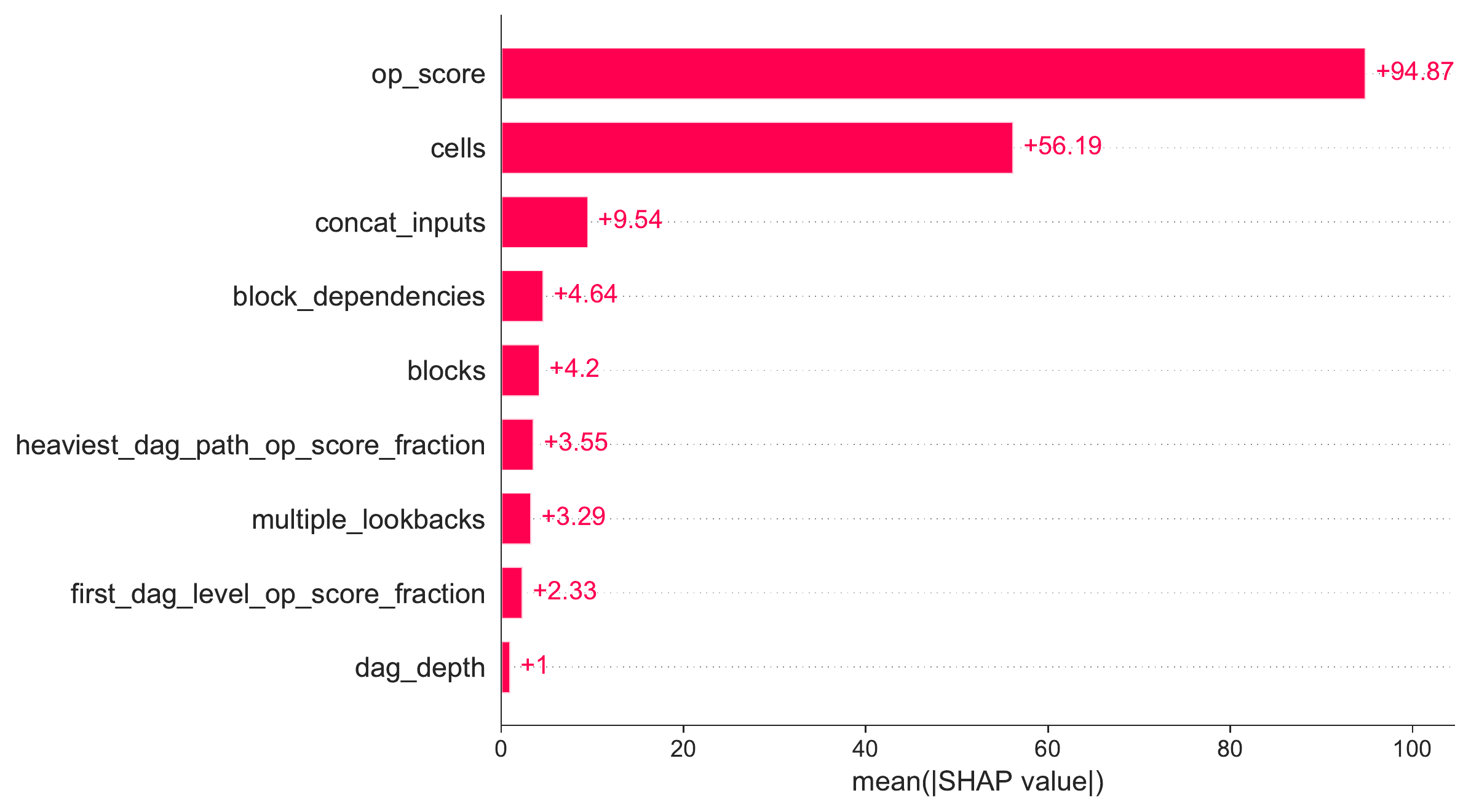}
	}
	\hfill
	\subfloat[SHAP impact on output per feature]{
	    \label{FIG:shap-beeswarm}
	    \includegraphics[width=0.52\textwidth]{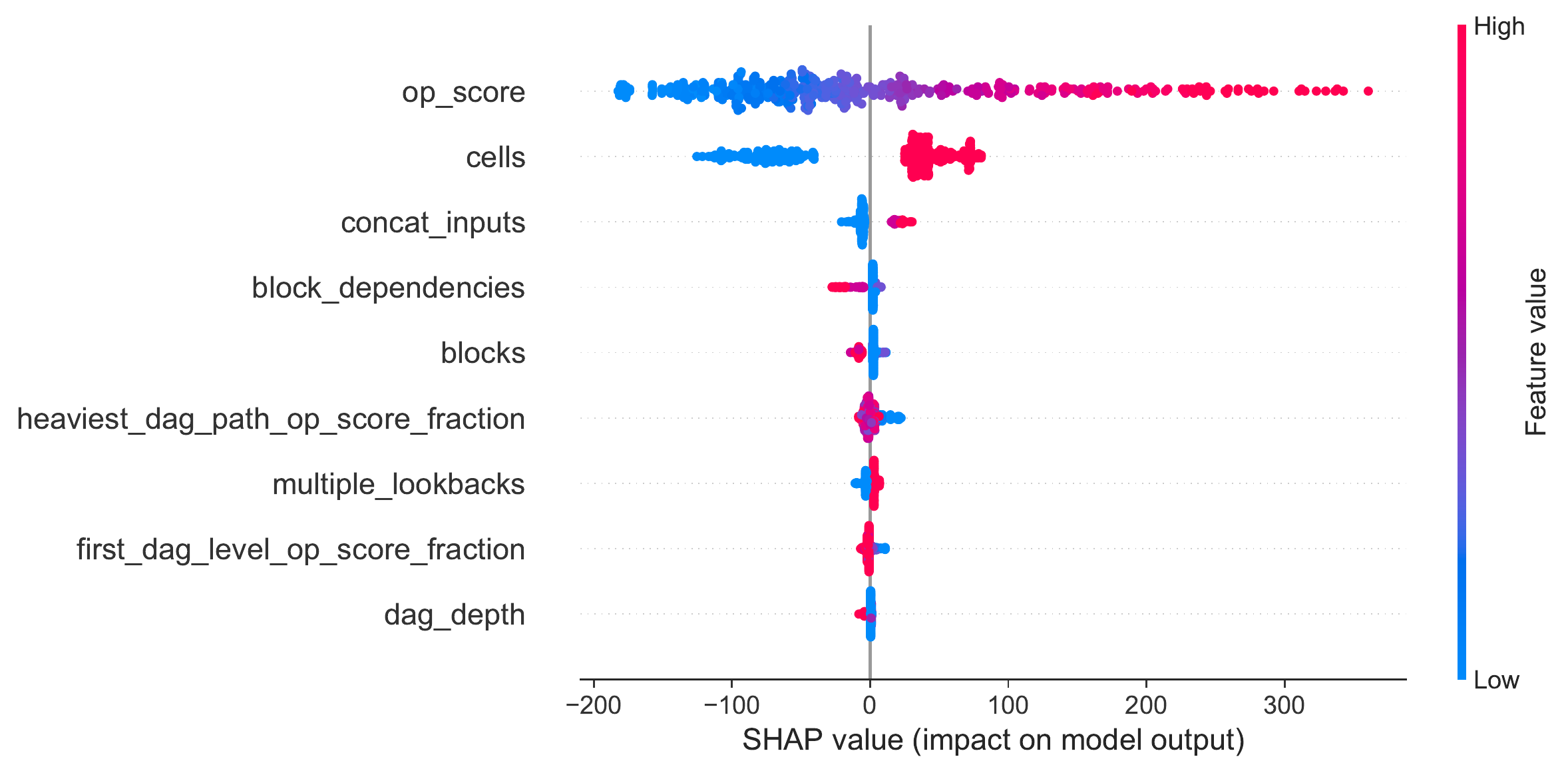}
	}
	\caption{Time predictor feature set analysis performed with the SHAP library. (a) An overview of how much the features are involved in the predictor outputs. (b) The actual numerical impact of the features on the training time predictions, in seconds.}
	\label{FIG:shap-analysis}
\end{figure}

POPNASv3 integrates the SHAP~\cite{SHAP} library, which we have used to perform some feature analysis and check which features are more relevant for the time changes between the architectures.
Figure~\ref{FIG:shap-analysis} reports the plots about the importance and impact on the final output of each feature.
The most important features for estimating the required training time of a cell are as expected the OP score, which is the sum of the dynamic reindex of all operators interconnected in the cell DAG, and the number of cells stacked in the final architecture.
The search configuration fixes the maximum number of cells stacked in the macro architecture, however, the actual amount of cells stacked can be reduced if the cell specification never uses the sequential lookback, i.e. -1 input value, so that some cells are not connected to the final output of the architecture DAG.

The number of unused blocks follows the two features previously mentioned, and this is to be expected since it involves an internal structural change in the cell.
When multiple block outputs are not used by other blocks, they are concatenated together and followed by a pointwise convolution to adapt the number of filters.
The pointwise convolution performed MAdds scale with the number of filters of the concatenation output, involving a time overhead proportional to the number of concatenated tensors.
Interestingly, features such as block dependencies, DAG depth and the percentage of OP score involved in the heaviest cell DAG path are not as relevant, although they should give a measure of how parallelizable the execution of the neural network layers is on the training device.
This observation could change with different frameworks, implementations and hardware.

\subsection{Search Strategy} \label{SEC:search-strategy}%

POPNASv3 starts the search from the simplest architectures of the search space and progressively builds more complex architectures by adding blocks to those previously sampled.
This type of search procedure, which alternates between fitting models and exploiting their results to explore unseen regions of the search space, is referred to in the literature as sequential model-based optimization~\cite{smbo}.
In our work, time and accuracy predictors learn from previous network results which blocks are more suited to expand the currently evaluated architectures, progressively increasing the model complexity until a target amount of blocks is reached.
A complete overview of the search strategy is provided in Algorithm~\ref{ALG:popnasv2_search}.

The procedure starts with the training of the \emph{empty cell}, which is the architecture composed only of the GAP and Softmax layers.
The importance of the empty cell lies in the fact that it captures elements shared by all architectures of the search space.
The reached accuracy $ a_{0} $ and required training time $ t_{0} $ can therefore be fed to predictors to calibrate the bias and improve the estimates.
Taking the training time $ t_{0} $ as an example, it measures the time required to execute the final layers common to all architectures, plus the time overhead due to data preprocessing and data augmentation procedures.
After training the empty cell, the algorithm bootstraps the search by training all architectures with a single block included in the search space.
The accuracy $ \mathcal{A}_{1} $ and time $ \mathcal{T}_{1} $ reached by these architectures are collected to train the respective predictors.
While the accuracy predictor can directly process the cell specifications, i.e. lists of block encodings, the time predictor requires the extrapolation of a designed set of features from the cell structure.
A \emph{dynamic reindex map} is generated from $ t_{0} $ and $ T_{1} $, providing a numerical measure of the time complexity of each operator $ o \in \mathcal{O} $.
More details about the time features and the dynamic reindex formula are given in Section~\ref{SEC:time-predictor}.

\begin{algorithm}[t]
	\caption{POPNASv3 search strategy}
	\label{ALG:popnasv2_search}
	\begin{algorithmic}[1]
		\REQUIRE $B$ (max num of blocks), $ E $ (epochs), $ K $ (beam size), $ J $ (exploration beam size),\\ \textit{CNN-hp} (networks hyperparameters), \textit{dataset}.
		\STATE $ S_{0} = \text{empty cell}$
		\STATE $ \mathcal{A}_{0}, \mathcal{T}_{0} = \text{train-cells}(S_{0}, E, \text{CNN-hp}, \text{dataset})$ 
		\STATE $ S_{1} = \mathcal{B}_{1} $ 
		\STATE $ \mathcal{A}_{1}, \mathcal{T}_{1} = \text{train-cells}(S_{1}, E, \text{CNN-hp}, \text{dataset})$
		\STATE $ \text{dynamic-reindex-map} = \text{compute-dynamic-reindex-map}(\mathcal{T}_{0}, \mathcal{T}_{1}) $
		\FOR{$ b = 2 : B $}
		    \STATE $ \mathcal{P}_{acc} = \text{fit}(\mathcal{A}_{0 \to b-1}, S_{0 \to b-1}) $
			\STATE $ \mathcal{F}_{b-1} = \text{extract-time-features}(S_{b-1}, \text{dynamic-reindex-map}) $
			\STATE $ \mathcal{P}_{time} = \text{fit}(\mathcal{T}_{0 \to b-1}, \mathcal{F}_{0 \to b-1}) $
			\STATE $ S_{b}' = \text{expand-cells}(S_{b-1}) $
			\STATE $ \mathcal{A}_{b}' = \text{predict}(S_{b}', \mathcal{P}_{acc}) $
			\STATE $ \mathcal{T}_{b}' = \text{predict}(S_{b}', \mathcal{P}_{time}) $
			\STATE $ S_{b} = \text{build-pareto-front}(S_{b}', \mathcal{A}_{b}', \mathcal{T}_{b}', K) $
			\STATE $ \tilde{\mathcal{O}}, \tilde{\mathcal{I}_{b}} = \text{build-exploration-sets}(S_{b}) $
			\IF{$ |\tilde{\mathcal{O}}| > 0 \lor |\tilde{\mathcal{I}_{b}}| > 0 $}
				\STATE $ S_{b, exp} = \text{build-exploration-pareto-front}(S_{b}', \mathcal{A}_{b}', \mathcal{T}_{b}', S_{b}, J) $
			\ELSE
				\STATE $ S_{b, exp} = \{\} $
			\ENDIF
			\STATE $ \mathcal{A}_{b}, \mathcal{T}_{b} = \text{train-cells}(S_{b} \cup S_{b, exp}, E, \text{CNN-hp}, \text{dataset})$
		\ENDFOR
	\end{algorithmic}
\end{algorithm}

From this point on, the search proceeds iteratively, analyzing the previous results at each step to converge to the optimal region of the search space and expand the architectures in that direction with an additional block.
At each iteration, the time and accuracy predictors are fitted on the data of all previously trained architectures.
After that, the algorithm lists all possible network expansions $ \mathcal{S}'_{b} $ of the architectures trained in the previous step ($ \mathcal{S}_{b-1} $), which are obtained by adding any single block to a cell specification.
During the expansion step, POPNASv3 forbids the generation of specular block encodings, which would lead to graph isomorphisms and so to computationally-equivalent neural networks.
Figure~\ref{FIG:block} illustrates the block isomorphism problem.
The identification of graph isomorphisms is critical to avoid wasting resources in training the same network multiple times, which would lead to a reduction in the variety of networks and an increase in search time.
The predictors are then used to estimate $ \hat{a}_{s}, \hat{t}_{s} $ of each architecture $ s \in \mathcal{S}'_{b} $, permitting the construction of a time-accuracy Pareto front $ \mathcal{S}_{b} $ with the architectures achieving the best tradeoff between the two metrics.
While building the Pareto front, POPNASv3 detects and prunes potential graph isomorphisms between candidate cells and the ones already inserted in the Pareto front.
Cells isomorphism detection is not performed during cell expansion since graph isomorphism is an NP-complete problem and it is computationally demanding to apply on sets of large cardinality.
Since the Pareto front is limited up to $ K $ elements, it is instead doable for such amount of networks.

In addition to the Pareto front architectures, an exploration step can potentially introduce new architectures to train, triggered only in case the predictors highly discourage the usage of some operators or inputs.
Going into the details, the algorithm counts how many times each $ o \in \mathcal{O} $ and $ i \in \mathcal{I}_{b} $ appear in the Pareto front networks and it inserts the operator and input values used respectively less than $ (5|\mathcal{O}|)^{-1} $ and $ (5|\mathcal{I}|)^{-1} $ into the exploration operator set $ \tilde{\mathcal{O}} $ and exploration input set $ \tilde{\mathcal{I}}_{b} $.
If the exploration sets are not empty, an additional Pareto front is constructed with only those architectures that contain at least a fixed number of inputs and operators from the exploration sets, eliminating the architectures already selected in the standard Pareto front and those that are isomorphic.
The iteration ends with the training of all architectures belonging to the Pareto front or exploration Pareto front, collecting as usual their results to train the predictors in future iterations.
POPNASv3 terminates when the architectures have been expanded to the target amount of blocks $ B $ defined in the experiment configuration.

\subsection{Post-Search Procedures}
The architectures sampled during the search are trained only for a small number of epochs, so they do not converge to their best results and require additional training.
Two extra steps, which we refer to as \emph{model selection} and \emph{final training}, are performed after the search procedure to determine the best architecture after prolonged training and maximize its performance.
While the search procedure focuses only on the micro-architecture, i.e. finding the best cell structures, the model selection can tune the macro-architecture by testing different configurations.
The final training tunes extensively the weights of the best architecture found during model selection.
A deployment-ready neural network model can therefore be produced by sequentially executing these two procedures after the search completes.


The search accuracy results, even if found on a short training, are indicative of the potentiality of the sampled networks and their generalization capabilities, so we simply select the top-K architectures resulting from the search for the model selection process.
The model selection step consists in training the candidate architectures for a significant amount of epochs, using the same training and validation dataset splits utilized during the search, so that we can determine the best architecture for the task.
An analogous procedure is also performed in~\cite{zoph2018learning}, but in POPNASv3 we limit it to the top-5 architectures by default, as the Pareto optimization ranks the architectures and makes differences between cells more evident.
In case particular deployment constraints exist regarding metrics like params and inference time, it is possible to exclude architectures not satisfying the constraints, as these metrics are computed and logged during the search procedure.
During model selection, we introduce additional generalization techniques, like scheduled drop path, introduced in~\cite{zoph2018learning} as a modification of the drop path~\cite{fractalnet} technique, whose drop rate scales with the number of iterations, and cutout~\cite{Cutout} data augmentation.
These techniques are generally not advised for the proxy training performed during the search, since they slow down the learning of the architectures, but are fundamental to avoid overfitting when training for a large number of epochs.

The model selection procedure is illustrated in Algorithm~\ref{ALG:model-selection}.
Model selection can optionally tune the macro architecture structure, i.e. the number of motifs $ M $, the number of normal cells $ N $ stacked in each motif, and the number of starting filters $ F $.
These parameters are fixed during the search and not optimized to avoid an exponential increase of the search space cardinality.
Architectures are trained using the macro-structure configuration provided during the search, then, if a range of parameters is provided as argument, the macro-architecture of these cells is tuned using grid search on sets of modifiers.
The modifiers are defined by the algorithm as the number of motifs and normal cells per motif to add compared to the search configuration, and a multiplier for the starting filters.
For each $ (\text{cell-encoding}, M, N) $ triplet, the algorithm trains the architecture with the largest amount of filters $ F $ fitting the defined range of parameters; in case no $ F $ value satisfies the parameters constraint, the triplet is simply discarded.

\begin{algorithm}[t]
	\caption{Model selection}
	\label{ALG:model-selection}
	\begin{algorithmic}[1]
		\REQUIRE \textit{search-results}, \textit{search-hp}, \textit{model-selection-hp}, \textit{dataset}, $ K $ (top networks to consider), $ P_{min} $, $ P_{max} $.
		\STATE $ M, N, F = \text{get-original-macro-config}(\text{search-hp})$
		\STATE $ \text{training-config} = \text{merge-configs}(\text{search-hp}, \text{model-selection-hp})$ 
		\STATE $ \mathcal{C} = \text{get-top-k-cells}(\text{search-results}, K) $
		\STATE $ M_{mod} = [0, 1], N_{mod} = [0, 1, 2], F_{mod} = [0.85, 1, 1.5, 1.75, 2] $
		\FOR{$ c \in \mathcal{C} $}
		    \STATE $ a_{c,M,N,F} = \text{train-architecture}(c, M, N, F, \text{training-config}, \text{dataset}) $
			\FOR{$ (m_{mod}, n_{mod}) \in (M_{mod} \times N_{mod}) $}
			    \STATE $ M' = M + m_{mod}, N' = N + n_{mod} $
			    \STATE $ F' = \text{max}(\{ f' = F \cdot f_{mod} | f_{mod} \in F_{mod} \land P_{min} < \text{get-params}(c, M', N', f') < P_{max} \}) $
			    \STATE $ a_{c,M',N',F'} = \text{train-architecture}(c, M', N', F', \text{training-config}, \text{dataset}) $
			\ENDFOR
		\ENDFOR
		\RETURN $ (c, m, n, f) $ with max$(a_{c, m, n, f})$
	\end{algorithmic}
\end{algorithm}

After performing model selection, the best architecture configuration is re-trained using all the available data for an even larger amount of epochs, aiming for the best deployment performances.
This means that in the final training the dataset is no more split into training and validation, so this last step should be performed only if the architecture generalizes well during the model selection training, otherwise there is a risk of overfitting.
Except for the number of epochs and the data split, the hyperparameters are set exactly as in the model selection phase.
When the training terminates, the architecture is evaluated on the test set to get the final results.

\section{Results}
In this section, we describe the experiments performed to assess POPNASv3 performance and compare the results with PNAS, the method out of which our search strategy is designed. 
We have performed the experiments on multiple image classification~(IC) and time series classification~(TSC) datasets commonly used in the literature, checking the algorithm robustness for different input sizes, channels, number of classes and domains.
The image classification datasets and their main characteristics are presented in Table~\ref{TAB:ic-datasets}.
Regarding TSC, we first considered the four datasets listed in Table~\ref{TAB:tsc-datasets} to perform preliminary tests, and we later extended the experiments to a larger selection of datasets.
In this second set of experiments, we selected all datasets from the UCR and UEA archives with at least 360 training samples, as neural networks could easily overfit with fewer data and the accuracy predictions could become noisy with fewer samples.
In these latter cases, we argue that NAS is not an adequate solution and that targeted training pipelines, together with domain-specific data augmentation techniques, produce better models.

Since POPNASv3 is a direct expansion of the PNAS search procedure using an improved model structure, we performed the image classification experiments also on POPNASv3 without Pareto optimization and the architecture modifications introduced by this work, simulating a PNAS execution on the same search space. Indeed, the original PNAS search algorithm code is not publicly available for a direct comparison.
Time series classification instead was not targeted by PNAS authors, therefore we only provide the results of POPNASv3 for these datasets.

\begin{table}[t]
	\caption{Datasets selected for image classification experiments. $^{*}$: EuroSAT provides only the training set, the test set is reserved with the last 20\% of samples, as done by EuroSAT authors themselves~\cite{helber2019eurosat}. }
	\begin{tabularx}{\linewidth}{@{} p{3.5cm} XXXXl @{}}
		\toprule
		Dataset & Train size & Test size & Input size & Channels & \# Classes \\
		\midrule
		CIFAR10~\cite{Krizhevsky09learningmultiple} & 50000 & 10000 & 32x32 & 3 (RGB) & 10 \\
		CIFAR100~\cite{Krizhevsky09learningmultiple} & 50000 & 10000 & 32x32 & 3 (RGB) & 100  \\
		Fashion MNIST~\cite{xiao2017fashionmnist} & 60000 & 10000 & 28x28 & 1 (Grayscale) & 10  \\
		EuroSAT$^{*}$~\cite{helber2019eurosat} & 21600 & 5400 & 64x64 & 3 (RGB) & 10 \\
		\bottomrule
	\end{tabularx}
	\label{TAB:ic-datasets}
\end{table}

\begin{table}[t]
	\caption{Datasets selected for the initial time series classification experiments.}
	\begin{tabularx}{\linewidth}{@{} p{3.5cm} XXXp{3cm}l @{}}
		\toprule
		Dataset & Train size & Test size & Length & Channels & \# Classes \\
		\midrule
		ElectricDevices~\cite{UCRArchive2018} & 8926 & 7711 & 96 & 1 (Univariate) & 7 \\
		FordA~\cite{UCRArchive2018} & 3601 & 1320 & 500 & 1 (Univariate) & 2 \\
        FaceDetection~\cite{UEAArchive2018} & 5890 & 3524 & 62 & 144 (Multivariate) & 2 \\
		LSST~\cite{UEAArchive2018} & 2459 & 2466 & 36 & 6 (Multivariate) & 14 \\
		\bottomrule
	\end{tabularx}
	\label{TAB:tsc-datasets}
\end{table}

\subsection{Experiments Configuration}

POPNASv3 models are implemented with the Keras API provided in TensorFlow 2.7.3~\cite{tensorflow2015-whitepaper}. All experiments have been performed on a single NVIDIA A100 GPU, partitioned using MIG configuration 3g.40gb, while model selection and final training have been carried out on the whole A100 without partitions.

We use a similar configuration for all experiments, adapting mainly the search space and the target amount of epochs in post-search procedures between image classification and time series classification tasks.
Table~\ref{TAB:hyperparameters} provides an overview of the main hyperparameters used in each step of the algorithm.
The operator set $ \mathcal{O} $ is adapted based on the input type.
For all image classification tasks, we use $ \mathcal{O} $ = \{ identity, 3x3 dconv, 5x5 dconv, 7x7 dconv, 1x3-3x1 conv, 1x5-5x1 conv, 1x7-7x1 conv, 1x1 conv, 3x3 conv, 5x5 conv, 2x2 maxpool, 2x2 avgpool \}; instead for time series classification we use $ \mathcal{O} $ = \{ identity, 7 dconv, 13 dconv, 21 dconv, 7 conv, 13 conv, 21 conv, 7:2dr conv, 7:4dr conv, 2 maxpool, 2 avgpool, lstm, gru \}.
The maximum lookback distance is set to 2 for all experiments.
The search continues until the cells are expanded to $ B=5 $ blocks, retaining a maximum of $ K=128 $ architectures in each step, plus up to $ J=16 $ additional architectures in case the exploration step is triggered.
During the search, the macro-structure is fixed for all models, building architectures with $ M=3 $ motifs, $ N=2 $ normal cells per motif and starting filters $ F=24 $.
The neural network models are trained for $ E=21 $ epochs, using AdamW optimizer with cosine decay restart scheduler for adapting the learning rate and weight decay during the training.
The learning rate is initially set to 0.01 and the weight decay to $ 5e^{-4} $ in IC tasks, while for TSC is increased to $ 1e^{-3} $.
This configuration allows fast convergence with minimal time investment, providing good estimations of the capabilities of each sampled architecture.
IC tasks use random shifting and random horizontal flip for data augmentation, while TSC datasets are just preprocessed by z-normalizing the data.

\begin{table}[t]
	\caption{Summary of hyperparameters used during the neural architecture search procedure and the post-search steps. $^{\dagger}$: image classification only. $^{\ddagger}$: time series classification only.}
	\begin{tabularx}{\linewidth}{@{} p{4cm} XXl @{}}
		\toprule
		Hyperparameter & Neural architecture search & Model selection & Last training \\
		\midrule
		epochs & 21 & 200 $^{\dagger}$, 100 $^{\ddagger}$ & 600 $^{\dagger}$, 200 $^{\ddagger}$ \\
		learning rate & 0.01 & 0.01 & 0.01 \\
		weight decay & 5$e^{-4}$ $^{\dagger}$, 1$e^{-3}$ $^{\ddagger}$ & 5$e^{-4}$ $^{\dagger}$, 1$e^{-3}$ $^{\ddagger}$ & 5$e^{-4}$ $^{\dagger}$, 1$e^{-3}$ $^{\ddagger}$ \\
		optimizer & AdamW & AdamW & AdamW \\
		scheduler & cosine decay restart & cosine decay & cosine decay \\
		drop path rate & 0.2 $^{\ddagger}$ & 0.4 $^{\dagger}$, 0.6 $^{\ddagger}$ & 0.4 $^{\dagger}$, 0.6 $^{\ddagger}$ \\
		label smoothing &  & 0.1 & 0.1 \\
		\midrule
		M & 3 & tuned & from model selection \\
		N & 2 & tuned & from model selection \\
		filters & 24 & tuned & from model selection \\
		secondary exit &  & \checkmark & \checkmark \\
		\midrule
		validation set size & 10\% & 10\% &  \\
		data augmentation $^{\dagger}$ & \checkmark & \checkmark & \checkmark \\
		cutout $^{\dagger}$ &  & \checkmark & \checkmark \\
		z-normalization $^{\ddagger}$ & \checkmark & \checkmark & \checkmark \\
		\bottomrule
	\end{tabularx}
	\label{TAB:hyperparameters}
\end{table}

After the search, the top-5 architectures are tuned in the model selection phase, using as modifiers: $ M_{mod} = [0, 1] $, $ N_{mod} = [0, 1, 2] $ and $ F_{mod} = [0.85, 1, 1.5, 1.75, 2] $.
For both the model selection and last training phases we use as a basis the hyperparameters configuration used during the search, overriding some parameters to perform a prolonged training.
The networks found on IC datasets are trained for 200 epochs during the model selection phase and for 600 epochs during the final training of the best architecture, while in TSC tasks the numbers of epochs are respectively reduced to 100 and 200 since they converge much faster.
The scheduler is changed to cosine decay, scheduled drop path is enabled with a rate of 0.4 in IC and 0.6 in TSC tasks, and label smoothing~\cite{LabelSmoothingHelp} is set in the loss function, with value 0.1.
A secondary output is added at 2/3 of the total architecture length to improve the generalization, the loss weights are set to 0.75 for the main output and 0.25 for the secondary output.
For image classification tasks, we also enable random cutout~\cite{Cutout} of size (8, 8), to further improve generalization.

The accuracy predictor is an ensemble of 5 LSTM-based models with attention having the structure defined in Section~\ref{SEC:acc-predictor}.
The ensemble is trained using K-folds, with $ K = 5 $, where each model is fit on a different fold.
The models share the same hyperparameters, using a learning rate of 0.004 with Adam~\cite{kingma2017adam} optimizer and no scheduler.
Weight regularization is applied with a factor $ 1e^{-5} $.

The Catboost regressor used as the time predictor performs random-search hyperparameter optimization at runtime, looking for the best parameters within this search space: \\
\begin{tabularx}{\linewidth}{l l}
    learning\_rate: & uniform(0.02, 0.2) \\
    depth: & randint(3, 7) \\
    l2\_leaf\_reg: & uniform(0.1, 5) \\
    random\_strength: & uniform(0.3, 3) \\
    bagging\_temperature: & uniform(0.3, 3) \\
\end{tabularx}
The training procedure uses K-folds, with $ K = 5 $, training each model for 2500 iterations and using early-stopping with patience set to 50 iterations.

\subsection{Predictor Results and Pareto Front Analysis}

In this section, we evaluate the performance of POPNASv3 predictors for both image classification and time series classification tasks.
The construction of the Pareto front depends on the correct ranking of the candidate architectures, so it is fundamental to evaluate the ranking quality of the predictions in each step.
To analyze the ranking and the accuracy we use respectively the Spearman's rank correlation coefficient and the mean average percentage error~(MAPE).
The accuracy predictor results are provided in Table~\ref{TAB:accuracy_predictors_results}, while the time predictors results are provided in Table~\ref{TAB:time_predictors_results}.
We also performed some experiments on different hardware configurations, to check if the predictors can adapt to the training device, making the procedure robust to various distribution strategies.
The results of the hardware experiments are reported in Appendix~\ref{SEC:Apx-B-hw-predictors}.

\begin{table}[t]
	\caption{The results of POPNASv3 accuracy predictor for each evaluated dataset.}
	\begin{tabularx}{\linewidth}{@{} p{3cm} XXXX @{\hskip 1cm} XXXl @{}}
		\toprule
		\multirow{2}{*}{Dataset} & \multicolumn{4}{c}{MAPE(\%)} & \multicolumn{4}{c}{Spearman($ \rho $)} \\
		\cmidrule{2-5} \cmidrule{6-9}
		& b=2 & b=3 & b=4 & b=5 & b=2 & b=3 & b=4 & b=5 \\
		\midrule
		CIFAR10 & 1.967 & 1.304 & 0.927 & 0.777 & 0.888 & 0.832 & 0.882 & 0.960 \\
		CIFAR100 & 13.492 & 1.635 & 2.354 & 2.766 & 0.909 & 0.820 & 0.947 & 0.945 \\
		Fashion MNIST & 1.743 & 0.779 & 1.063 & 0.940 & 0.663 & 0.531 & 0.773 & 0.935 \\
		EuroSAT & 1.325 & 0.874 & 0.615 & 0.597 & 0.605 & 0.840 & 0.810 & 0.872 \\
		\midrule
		ElectricDevices & 1.934 & 1.321 & 1.144 & 0.595 & 0.830 & 0.945 & 0.818 & 0.817 \\
		FordA & 2.842 & 0.717 & 0.652 & 0.822 & 0.633 & 0.568 & 0.524 & 0.688 \\
		FaceDetection & 7.375 & 2.943 & 2.743 & 1.952 & 0.750 & 0.551 & 0.247 & 0.477 \\
        LSST & 7.504 & 9.208 & 15.393 & 14.716 & 0.827 & 0.822 & 0.736 & 0.775 \\
		\bottomrule
	\end{tabularx}
	\label{TAB:accuracy_predictors_results}
\end{table}

\begin{table}[t]
	\caption{The results of POPNASv3 time predictor for each evaluated dataset.}
	\begin{tabularx}{\linewidth}{@{} p{3cm} XXXX @{\hskip 1cm} XXXl @{}}
		\toprule
		\multirow{2}{*}{Dataset} & \multicolumn{4}{c}{MAPE(\%)} & \multicolumn{4}{c}{Spearman($ \rho $)} \\
		\cmidrule{2-5} \cmidrule{6-9}
		& b=2 & b=3 & b=4 & b=5 & b=2 & b=3 & b=4 & b=5 \\
		\midrule
		CIFAR10 & 17.219 & 10.134 & 5.487 & 8.709 & 0.960 & 0.979 & 0.991 & 0.979 \\
		CIFAR100 & 21.830 & 6.459 & 6.334 & 6.599 & 0.982 & 0.995 & 0.990 & 0.994 \\
		Fashion MNIST & 23.912 & 19.755 & 11.614 & 11.512 & 0.965 & 0.962 & 0.964 & 0.922 \\
		EuroSAT & 17.852 & 7.815 & 5.839 & 4.707 & 0.976 & 0.995 & 0.996 & 0.994 \\
		\midrule
		ElectricDevices & 27.792 & 12.390 & 11.198 & 6.377 & 0.920 & 0.970 & 0.979 & 0.987 \\
		FordA & 26.059 & 14.740 & 6.802 & 3.827 & 0.879 & 0.958 & 0.896 & 0.919 \\
		FaceDetection & 13.781 & 11.214 & 7.479 & 13.054 & 0.646 & 0.832 & 0.922 & 0.947 \\
        LSST & 16.508 & 22.587 & 33.984 & 10.300 & 0.717 & 0.966 & 0.836 & 0.940 \\
		\bottomrule
	\end{tabularx}
	\label{TAB:time_predictors_results}
\end{table}

The accuracy predictions made by the LSTM ensemble have generally small errors, with some exceptions.
We notice that the inaccuracies happen especially during $ b=2 $ step in datasets difficult to fit in a short amount of epochs; in these cases, the predictor tends to overestimate the gain obtained by adding the second block, but this behavior is fixed in the later iterations.
The ranking provided by the predictor is quite accurate in image classification tasks, contributing greatly to the Pareto front accuracy.
Concerning time series classification tasks, the results tend to be noisier, but this behavior is to be expected given the smaller amount of available data and the lack of generic data augmentation techniques.

Regarding time predictions, the MAPE is significantly higher in the $ b=2 $ step.
The error is due to features related to structural changes in multi-block cells, which are constant for architectures with a single block, therefore it is impossible for the time predictor to learn the importance of these features without pretraining.
Although this is a limitation of our feature set, the error is a bias common to most architectures, thus it does not affect the ranking.
The MAPE tends to decrease in later steps, since the predictor can learn these behaviors specific to more complex cells.
Nevertheless, the ranking in time predictions is optimal, with most Spearman coefficients tending to 1.
The ranking quality offsets the suboptimal MAPE in the early stages, as Pareto front correctness is the key to increasing search efficiency without discarding potential optimal solutions.

\begin{figure}[t]
	\centering
	\subfloat[Pareto front computed from predictions]{
	    \label{FIG:only-predictions-pareto}
	    \includegraphics[width=0.48\textwidth]{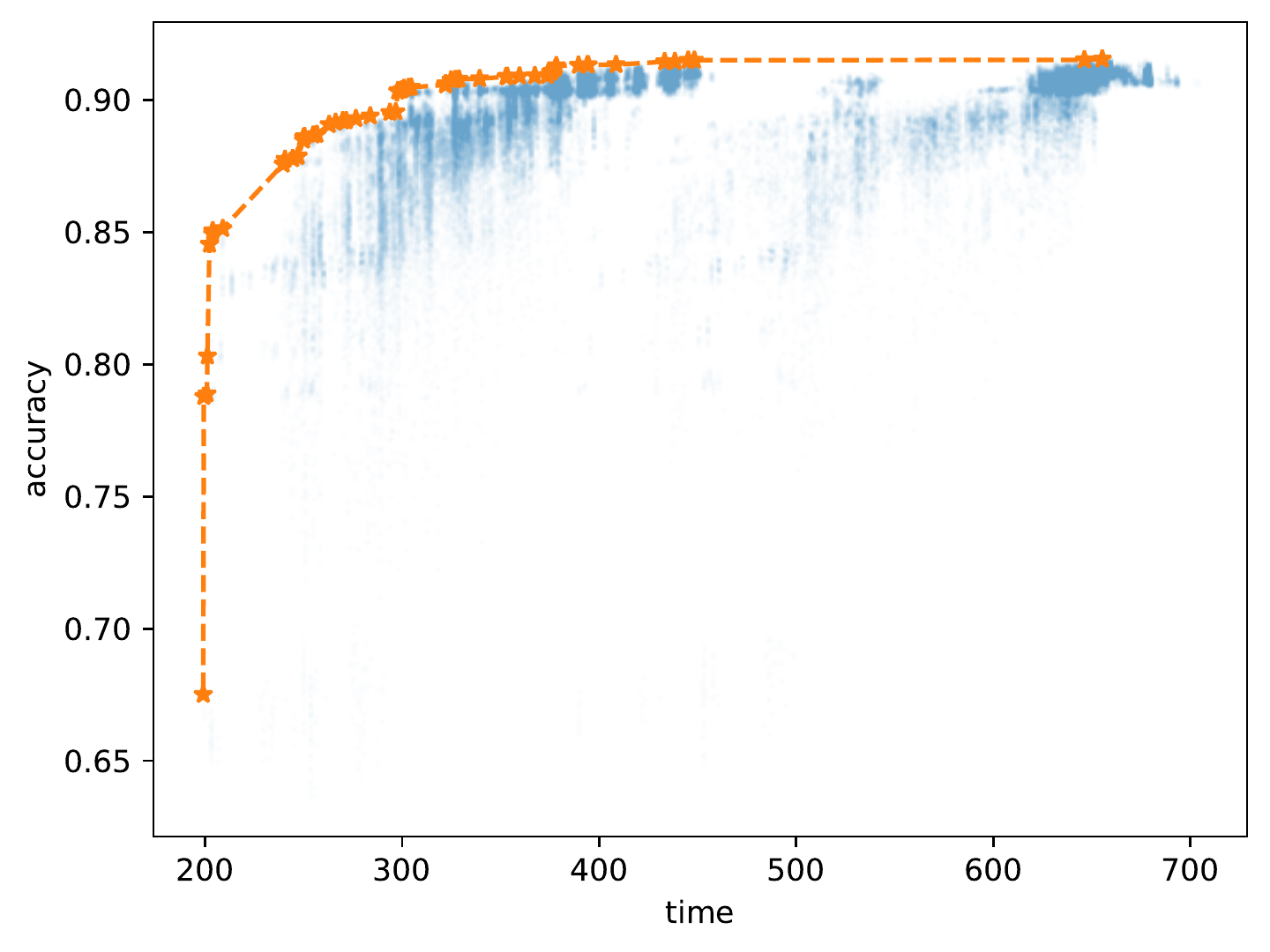}
	}
	\hfill
	\subfloat[Predicted Pareto front vs real results]{
	    \label{FIG:predictions-vs-real-pareto}
	    \includegraphics[width=0.46\textwidth]{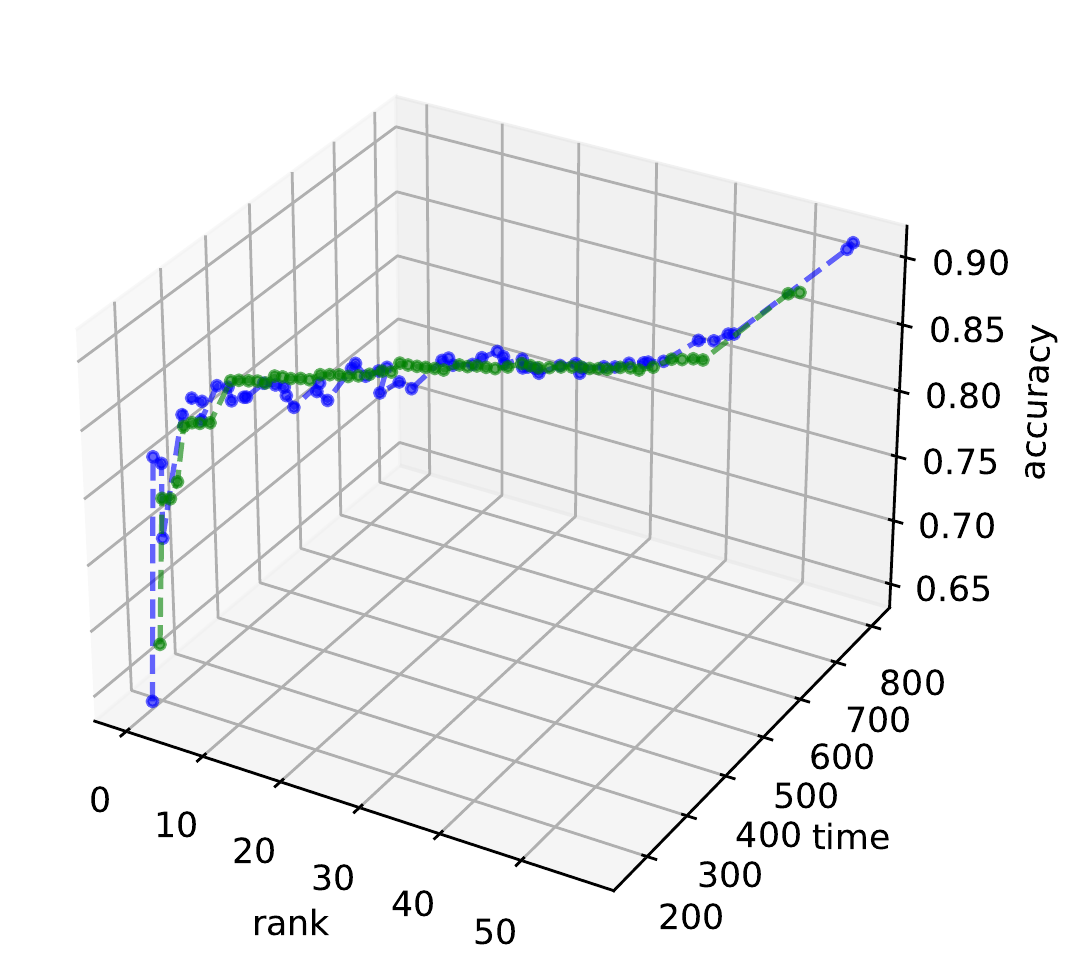}
	}
	\caption{Pareto plots extrapolated from CIFAR10 experiment, on architectures with cells containing 5 blocks.
    (a) The scatter plot of the estimates made by predictors, for each network candidate evaluated during the expansion from 4 to 5 blocks. Each blue semi-transparent pixel represents a different cell specification, opaque point clouds denote a high number of architectures having similar results.
    POPNASv3 selects for training only the architectures belonging to the Pareto front, here depicted with the orange star symbols.
    (b) A comparison of the real and estimated Pareto fronts. The blue line shows the real accuracy and training time retrieved after the proxy training, while the green line plots the estimated values indicated by the predictors.}
	\label{FIG:pareto-analysis}
\end{figure}

The plots shown in Figure~\ref{FIG:pareto-analysis} give an overview of how the Pareto optimization is effective for the quick exploration of vast search spaces.
Even if each expansion step of POPNAS is restricted to select at most $ K $ architectures, the cardinality of the possible cell expansions is orders of magnitude higher than $ K $.
From Figure~\ref{FIG:only-predictions-pareto} it is clear that there is a high density of architectures with promising accuracy, but it is a waste of resources to consider them all as done in PNAS since they are often associated with larger training time and their estimated accuracy gap is not significantly better.
Furthermore, the time-accuracy Pareto front often contains a number of architectures lower than $ K $, resulting in fewer architectures to sample, which speed-ups the search even further.
Although more costly, sampling the top accuracy networks as done by PNAS is more error-proof for finding the most accurate cell, given the fact that the predictors could produce some outliers and noisy estimates.
In this regard, the POPNASv3 exploration step introduces in each iteration a few cells which utilize inputs and operators rarely selected in the Pareto front architectures.
The results of these peculiar cells help the predictor in selecting more diversified networks, in case their results compete with the ones selected in the Pareto front.
Therefore, the exploration step mitigates eventual noise in the predictions due to biases introduced by the simplest architectures and stabilizes the convergence to the optimal search space regions.

\subsection{Image Classification Results} \label{SEC:IC-results}%


For the image classification datasets, we have run the experiments on both POPNASv3 and PNAS, using the same hyperparameters configuration to make a fair comparison.
Besides missing the Pareto front and exploration step from the procedure, PNAS does not use residual units at cell outputs and reshapes the lookbacks with a pointwise before generating the next cell, in case their shape is not equal.
Other improvements introduced by POPNASv3 which do not alter the search strategy and macro architecture, like the use of Swish as the activation function and the detection of graph isomorphisms, are instead ported to our PNAS implementation.

Table~\ref{TAB:search-results-IC} shows the search results obtained with the two methods.
Considering the fact that POPNASv3 and PNAS bootstrap the search in the same way and they use the same search space in our experiments, the first 301 sampled cell specifications are equivalent in the two methods.
In the next computations, we exclude these architectures from the network count and search time to better estimate the effect of the Pareto optimality on search efficiency.
POPNASv3 trains on average 56\% fewer architectures than the target beam size $ K $, fixed in the configuration and always saturated by PNAS in each step given its search strategy definition.
The restricted cardinality of the Pareto front contributes greatly to the reduction of the search time, but the time-accuracy optimization also reduces the average training time of these networks, achieving far better speed-ups than the estimated 2x caused by architecture reduction.
From the results, POPNASv3 speed-ups the search procedure on average by a 4.8x factor when considering only $ b >= 2 $ steps, while on the whole search procedure the improvement is still a significant 3.4x factor.
Regarding the accuracy reached after the proxy training, the top architectures found by POPNASv3 and PNAS have no significant differences, except on the CIFAR100 dataset.
This latter gap is attributable to the residual connections added to the output of the POPNASv3 cells, which improve training convergence by facilitating gradient propagation, enhancing the accuracy in datasets where the proxy training results are far from the optimal ones.
Since POPNASv3 samples fewer networks, we check if our algorithm can find multiple solutions with similar performances.
The average accuracy of the 5 best architectures found on each dataset is close to the top accuracy, implying that the Pareto optimization does not introduce large accuracy gaps between the most promising architectures.
Post-search procedures can therefore train more extensively these architectures and check which configuration performs better at convergence, based on possible deployment constraints.

\begin{table}[t]
	\caption{The comparison between POPNASv3 and PNAS search performance over the evaluated IC datasets. The listed accuracies are the ones obtained on the validation set after the proxy training.
    The number of networks and the search time written in brackets refer to steps with $ b > 2 $, excluding the $ b <= 1 $ steps which are deterministic and equivalent between the two methods.
 }
	\centering
	\begin{tabularx}{\linewidth}{@{} ll YYYl @{}}
		\toprule
		Dataset & Method & \# Networks & Top Cell Accuracy & Top-5 Cells Accuracy & Search Time \\
		\midrule
		\multirow{2}{*}{CIFAR10} & POPNASv3 & $ \mathbf{520} $ (219) & 0.918 & 0.917 & \textbf{46h 48m} (29h 19m) \\
		& PNAS & 813 (512) & $ \mathbf{0.922} $ & $ \mathbf{0.920} $ & 176h 30m (158h 12m) \\
		\addlinespace[0.6em]
		\multirow{2}{*}{CIFAR100} & POPNASv3 & $ \mathbf{539} $ (238) & $ \mathbf{0.706} $ & $ \mathbf{0.701} $ & \textbf{55h 5m} (37h 49m) \\
		& PNAS & 813 (512) & 0.680 & 0.679 & 161h 47m (143h 7m)  \\
		\addlinespace[0.6em]
		\multirow{2}{*}{Fashion MNIST} & POPNASv3 & $ \mathbf{474} $ (173) & $ \mathbf{0.952} $ & $ \mathbf{0.950} $ & \textbf{44h 0m} (26h 3m) \\
		& PNAS & 813 (512) & 0.946 & 0.945 & 174h 53m (155h 23m) \\
		\addlinespace[0.6em]
		\multirow{2}{*}{EuroSAT} & POPNASv3 & $ \mathbf{570} $ (269) & $ \mathbf{0.968} $ & $ \mathbf{0.967} $ & \textbf{83h 53m} (53h 35m) \\
		& PNAS & 813 (512) & 0.967 & 0.964 & 246h 14m (217h 45m) \\
		\bottomrule
	\end{tabularx}
	\label{TAB:search-results-IC}
\end{table}

\begin{table}[t]
	\caption{The comparison between POPNASv3 and PNAS top-1 networks test performance over the evaluated IC datasets.}
	\begin{tabularx}{\linewidth}{@{} ll YYYYYY l @{}}
		\toprule
		Dataset & Method & B & M & N & F & Params & Accuracy & Training Time \\
		\midrule
		\multirow{2}{*}{CIFAR10} & POPNASv3 & 4 & 3 & 2 & 36 & 3.46M & $ \mathbf{0.961} $ & \textbf{5h 22m} \\
		& PNAS & 5 & 3 & 3 & 24 & $ \mathbf{3.25M} $ & $ \mathbf{0.961} $ & 11h 25m \\
		\addlinespace[0.6em]
		\multirow{2}{*}{CIFAR100} & POPNASv3 & 5 & 3 & 2 & 24 & 4.08M & $ \mathbf{0.796} $ & \textbf{5h 13m} \\
		& PNAS & 5 & 3 & 2 & 24 & $ \mathbf{4.00M} $ & 0.794 & 5h 24m \\
		\addlinespace[0.6em]
		\multirow{2}{*}{Fashion MNIST} & POPNASv3 & 2 & 3 & 2 & 36 & $ \mathbf{3.11M} $ & $ \mathbf{0.957} $ & \textbf{2h 38m} \\
		& PNAS & 4 & 3 & 2 & 24 & 3.14M & 0.956 & 6h 3m \\
		\addlinespace[0.6em]
		\multirow{2}{*}{EuroSAT} & POPNASv3 & 4 & 3 & 3 & 20 & 3.37M & 0.983 & \textbf{9h 41m} \\
		& PNAS & 5 & 3 & 2 & 24 & $ \mathbf{3.34M} $ & $ \mathbf{0.987} $ & 13h 35m \\
		\bottomrule
	\end{tabularx}
	\label{TAB:final_training_results-IC}
\end{table}

After the search, we performed model selection over the top-5 cells found, using a parameters range of $(2.8e^{6}, 3.5e^{6})$ for macro-architecture tuning, to facilitate comparison with similar architecture designed for mobile settings.
Finally, we trained to convergence the best cell and macro configuration combination retrieved during model selection.
The final training results and their macro-architecture settings are summarized in Table~\ref{TAB:final_training_results-IC}, while the cell structures are illustrated in Figure~\ref{FIG:IC-best-cells}.
The training time required by POPNASv3 architectures is always lower than PNAS, even if the amount of parameters is almost identical.
This result is not surprising, since the training time is optimized by POPNAS during the search but it is not considered by PNAS.
The overall training results of the entire search indicate that parameters and FLOPs have a strong correlation with the training time, but architectures with similar parameters and flops can actually have remarkably different training times.
Some operators, in particular depthwise-separable and spatial-separable convolutions, appears less optimized for GPU parallelization since they consist of two separate convolutions executed sequentially.
These types of convolutions are therefore slower to perform than normal convolutions, although they have fewer parameters and FLOPs for the same kernel size.
This behavior could change based on the training hardware, but it was consistent in all our experiments on A100 GPUs.
The average accuracy of these architectures is not significantly different, which means that POPNASv3 speed-ups are not counterbalanced by performance loss in deployment-ready models.

\subsection{Time Series Classification Results} \label{SEC:TSC-results}%

Time series classification experiments have been carried out only with the POPNASv3 search strategy, since PNAS was not meant to work on time series.
We first analyze the results on the datasets considered for the preliminary tests.
Table~\ref{TAB:search-results-TSC} lists the results of the search procedure.
Similarly to what we observed in image classification cases, POPNASv3 trains a limited amount of architectures by exploiting Pareto optimization, converging quickly to a specific search space region.
On average there is minimal variance between the accuracy of the top-5 cell specifications, meaning that POPNASv3 found multiple competitive architectures to be tuned in the model selection step.
The search time required for time classification experiments is almost an order of magnitude lower than image classification tasks, which is justified by the fact that these datasets have fewer samples and in general fewer data to process per sample.
In our opinion, POPNASv3 is a valuable technique for exploring different operator sets and architectures on time series classification tasks, given the appealing computation time to perform the experiments.

\begin{table}[t]
	\caption{POPNASv3 search performance over the evaluated TSC datasets. The listed accuracies are the ones obtained on the validation set after the proxy training.}
	\begin{tabularx}{\linewidth}{@{} l YYYl @{}}
		\toprule
		Dataset & \# Networks & Top Cell Accuracy & Top-5 Cells Accuracy & Search Time \\
		\midrule
		ElecticDevices & 574 & 0.919 & 0.917 & 17h 42m \\
		FordA & 439 & 0.970 & 0.968 & 7h 4m \\
        FaceDetection & 453 & 0.791 & 0.790 & 6h 50m \\
		LSST & 481 & 0.663 & 0.645 & 5h 15m \\
		\bottomrule
	\end{tabularx}
	\label{TAB:search-results-TSC}
\end{table}

Also for this task, we finalize the search process by performing model selection, using a parameter range of $(1e^{6}, 2e^{6})$, and training the best cell and macro-architecture combination to convergence.
The results are presented in Table~\ref{TAB:final_training_results-TSC}, while the cell specifications are illustrated in Figure~\ref{FIG:TSC-best-cells}.
From the accuracy results, it is clear that the best validation accuracies found over the 90\%-10\% train-validation split are over-optimistic of the actual capabilities of the networks.
Indeed, the accuracy found on the test sets after prolonged training is tendentially lower than the validation accuracy reached after the proxy training.
This behavior indicates that the models can converge even during the proxy training, but the validation accuracy could be misleading due to overfitting on the small sample size or strong correlations between training and validation samples, e.g. when the samples are windows, potentially overlapped, of the same signal.
Expanding the validation split could lead to more accurate estimations at search time, but reducing the training data in datasets with fewer samples could impact negatively the learning process.
An alternative approach is to perform k-fold, as done in~\cite{NAS-T}, at the cost of multiplying the search time by the number of folds employed.

Regardless of the validation-test discrepancy, the final accuracy results obtained on the default train-test split of these datasets are remarkable, competing with state-of-the-art algorithms endorsed by the TSC community such as InceptionTime~\cite{InceptionTime}, HIVE-COTEv2~\cite{HIVE-COTEv2} and ROCKET~\cite{ROCKET}.
The comparison with these algorithms is presented in Table~\ref{TAB:SOTA-comparison-TSC}.
POPNASv3 achieves the best accuracy on both the considered multivariate datasets, but while on FaceDetection
the gap is not significant, on LSST
we achieve far superior results compared to the other methods.
RNN operators seem to be the key to LSST
performance; in fact, the top 30\% cells sampled during the search use LSTM or GRU at least once as an operator, with GRU often outperforming LSTM.
Combining RNN cells with Conv1D operators seems therefore beneficial for improving model performance on some TSC problems.
In the literature, this type of architecture goes under the name of convolutional recurrent neural networks~(CRNN), and it has seen application in TSC papers like~\cite{CRNN-music-class, CRNN-ECG-class}.
However, these works usually structure the network to extract features using only convolutions and aggregate them into the final layers with the RNN units.
Applying the RNN units directly on the input sequences and within the architecture, as done in POPNASv3 cells, works well in our experiments, indicating that more flexible CRNN structures should also be considered in these tasks.

\begin{table}[t]
	\caption{POPNASv3 top-1 networks performance over the default test split of the evaluated TSC datasets.}
	\centering
	\begin{tabularx}{\linewidth}{@{} l YYYYYY l @{}}
		\toprule
		Dataset & B & M & N & F & Params & Accuracy & Training Time \\
		\midrule
		ElectricDevices & 4 & 3 & 4 & 24 & 1.41M & 0.753 & 37m 4s \\
		FordA & 4 & 4 & 2 & 20 & 1.85M & 0.958 & 6m 20s \\
        FaceDetection & 3 & 3 & 4 & 24 & 1.01M & 0.676 & 9m 32s \\
		LSST & 4 & 3 & 4 & 24 & 1.80M & 0.721 & 11m 40s \\
		\bottomrule
	\end{tabularx}
	\label{TAB:final_training_results-TSC}
\end{table}

\begin{table}[t]
	\caption{Accuracy comparison between POPNASv3 and state-of-the-art methods on TSC datasets, using the original train-test split.}
	\centering
	\begin{tabularx}{\linewidth}{@{} l YYYc @{}}
		\toprule
		Dataset & HIVE-COTEv2~\cite{HIVE-COTEv2} & InceptionTime~\cite{InceptionTime} & ROCKET~\cite{ROCKET} & POPNASv3 \\
		\midrule
        ElectricDevices & \textbf{0.758} & 0.722 & 0.726 & 0.753 \\
		FordA & 0.954 & \textbf{0.964} & 0.946 & 0.958 \\
        FaceDetection & 0.660 & \textendash & 0.644 & \textbf{0.676} \\
        LSST & 0.643 & 0.612 & 0.637 & \textbf{0.721} \\
		\bottomrule
	\end{tabularx}
	\label{TAB:SOTA-comparison-TSC}
\end{table}

Dilated convolutions perform well in all datasets except ElectricDevices, but this behavior can be explained by analyzing the samples contained in this dataset.
The ROCKET authors provided interesting insights on how dilated convolutions can be understood as a way to sample the signal at different frequencies.
This behavior can be advantageous for finding patterns over a wide receptive range on slow-evolving signals, and the contribution of these operators has been statistically shown to increase the average performance of the ROCKET classifier.
This is also the case for POPNASv3, but here the architectures are adapted to use the best operators for the data.
ElectricDevices contains data about electric consumption from household devices.
Multiple classes of devices have prolonged flat sections at 0 in the signal, assumably because they are unused and disconnected in those time windows, and are thus characterized by short peaks of different amplitudes.
Sampling the ElectricDevices dataset at lower frequencies means not detecting these short peaks, so POPNASv3 discards dilated convolutions and uses only non-dilated ones.

These considerations provide evidence of how using NAS on large search spaces with assorted operators could lead to interesting architectures for the given problem, in a complete black-box manner.  
We argue that NAS goal should not focus only on achieving the best possible accuracy for the task, but to design algorithms that have the freedom of finding unconventional architectures, which might transcend the current knowledge we have on neural network architecture engineering.

The top cells found on these datasets are shown in Figure~\ref{FIG:TSC-best-cells} of the Appendix A.
All the cells tend to structure the computation graph in multiple levels, processing some of the intermediate block results with other operators to rapidly increase the receptive field and aggregate the features extracted by previous layers.
This structural behavior is not observed in the image classification cells, where \q{flat} cells perform well just by applying all operators in parallel.
Furthermore, all TSC cells use only -2 as lookback and therefore fewer cells than the target amount defined in the POPNASv3 configuration, indicating that the eventual performance boost obtained by adding more cells is negligible compared to the increased computational cost.
The operators selected in these cells highly depend on the dataset characteristics, as we analyzed before, and they are interconnected without detectable patterns.

\subsubsection{Extended Experiments}
As previously mentioned, we extend the experiments on a larger selection of datasets, including in this second analysis also the four datasets used in the preliminary tests.
The complete results on all these datasets are presented in Appendix A Section~\ref{SEC:TSC-extended-apx}.
The base configuration used for these experiments is the one provided in Table~\ref{TAB:hyperparameters}, but the batch size and validation size are tuned for each dataset.
The fraction of training samples reserved for validation is adapted to have at least 60 samples in the validation split, reducing the variance of the results in smaller datasets without decreasing too much the training size.
The batch size is assigned based on the number of samples in the training split, as the maximum multiplier of 16 in the interval~[32, 128], so that each epoch processes at least 10 batches of data.
This adjustment strikes a balance between the training time and the convergence stability: larger batch sizes improve the training efficiency, but having too few weight update steps does not work well with the learning rate scheduler and scheduled drop path, potentially causing underfitting.

Table~\ref{TAB:ext-TSC-search} summarize POPNASv3 search results on the extended selection of datasets.
Most datasets terminate the search procedure in less than 10 hours, sampling between 400 and 500 architectures.
Concerning the mean validation accuracy of the 5 best cells found on each dataset, it is usually close to the top result, considering also that the validation size for most datasets is just a few dozen of samples.
The final architectures retrieved after model selection with params range $(1e^{6}, 3e^{6})$, listed in Table~\ref{TAB:ext-TSC-final}, are quite heterogeneous both in structure and operator usage, confirming the observations found during the preliminary tests.
In summary, 29 of the best architectures found on the 49 considered datasets use only skip connections as lookback, keeping a smaller amount of cells to stack, but they often reintroduce some of them by increasing $ M $ and $ N $ parameters during model selection.
Only two groups of datasets, namely (Adiac, EthanolLevel) and (FordB, EOGVerticalSignal, UWaveGestureLibraryZ), share the same cells, while other datasets have unique architectures.

\begin{figure}[t]
	\centering
	\includegraphics[width=0.9\textwidth]{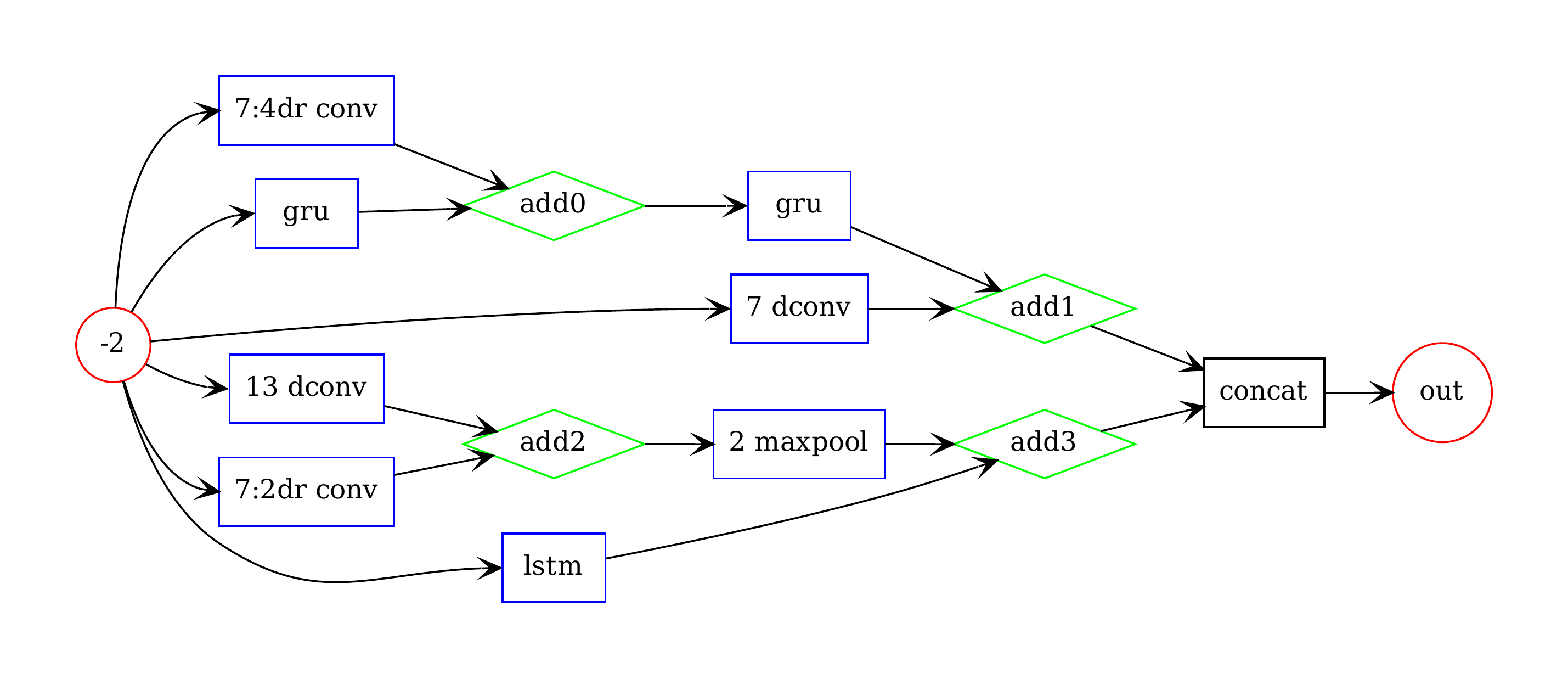}
	\caption{Cell structure found on PhonemeSpectra dataset.}
	\label{FIG:phoneme-cell}
\end{figure}

The comparison between the test accuracy found by POPNASv3 and the considered state-of-the-art methods of TSC literature are presented in Table~\ref{TAB:ext-TSC-comparison}.
On all four multivariate datasets, POPNASv3 is able to find better solutions than the other methods, indicating that the neural networks found by our algorithms are well suited to extract correlations from multiple time series.
Apart from LSST, which we already analyzed in previous experiments, also PhonemeSpectra dataset shows a significant improvement over the other considered models, achieving a 32\% relative accuracy gain over HIVE-COTEv2.
The architecture found on this complex dataset, reported in Figure~\ref{FIG:phoneme-cell}, is particularly interesting, since POPNASv3 mixes seven different operators in a convoluted DAG structure, interconnecting blocks together in multiple paths.
Multi-branch asymmetric architecture patterns are never explored in hand-crafted models, but these particular solutions could be interesting in different applications and are easier to explore with NAS.

\begin{figure}[t]
	\centering
	\includegraphics[width=\textwidth]{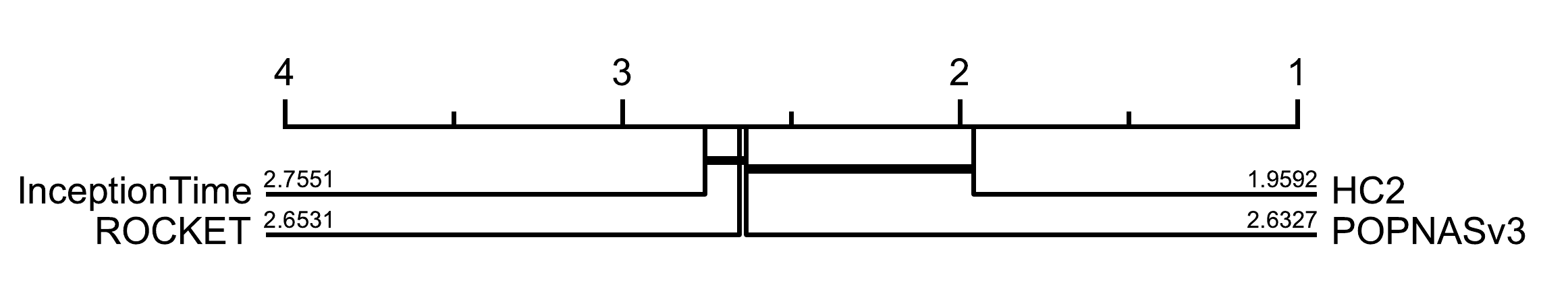}
	\caption{
     Critical difference diagram on the extended TSC dataset selection.
     The number associated with each method represents the mean rank obtained over the dataset selection.
     A thick black line, i.e. a clique, interconnects groups of datasets whose results are not considered statistically different when applying the Wilcoxon signed-rank test, meaning that the null hypothesis is rejected for each pair of methods belonging to the group.
     }
	\label{FIG:cd-diagram}
\end{figure}

On univariate datasets, POPNASv3 performance deteriorates, finding often architectures composed of a single block, which do not always compete with state-of-the-art solutions, even on the simplest datasets.
We aggregate the results on all datasets with the mean rank of each considered technique together with POPNASv3, represented with a critical difference diagram~\cite{JMLR:v7:demsar06a} using Wilcoxon signed-rank test with Holm alpha correction~\cite{JMLR:v9:garcia08a, JMLR:v17:benavoli16a}($\alpha = 0.05$), shown in Figure~\ref{FIG:cd-diagram}.\footnote{The figure has been generated with the open-source code provided in https://github.com/hfawaz/cd-diagram\cite{IsmailFawaz2018deep}.}
A thick black line, i.e. a clique, indicates a group of methods whose results are not statistically different.
POPNASv3 achieves the second-best mean rank, with similar results to the ones achieved by InceptionTime and ROCKET.
The Wilcoxon signed-rank test indicates that POPNASv3 results are statistically competitive with HIVE-COTEv2, even if HIVE-COTEv2 has a significantly better ranking on the whole selection.
As hinted before, POPNASv3 finds mono-block cells on about half the considered datasets, which is not inherently bad for simple-to-learn problems, but we suspect that the algorithm did not converge to optimal solutions in some cases.
The low amount of samples contained in these datasets can in fact hinder the search convergence, making the process prone to significant randomness.
The validation size on small datasets is not sufficient for properly estimating the quality of a network: when only dozens of samples are available for validation, retraining the same architecture can give quite different accuracy results.
The uncertainty of these measurements affects heavily the accuracy predictor, which becomes unable to distinguish architectures inside the uncertainty range, reflecting negatively on the Pareto front extracted by the algorithm.
This is a problem common to all machine learning techniques, but it is particularly hard to address in end-to-end solutions like NAS and we recognize that POPNASv3 is not well suited for general usage on small datasets.
On the datasets with more available samples and on multivariate datasets POPNASv3 produced architectures are instead peculiar and have competitive results, even significantly better on multivariate cases.

\section{Conclusion}
In this article, we presented POPNASv3, a sequential model-based optimization method that can work with potentially unlimited operator sets, while keeping competitive search time and accuracy results with other state-of-the-art methods.
Starting the search by training the simplest models allows the algorithm to identify the most promising operators and their training efficiency.
POPNASv3 exploits this information to progressively expand the architectures, considering a time-accuracy tradeoff that drastically speeds up the search by preserving only a limited set of peculiar architectures per iteration.
Our work can address image and time series classification tasks without modifications, except for the configuration file.
We show therefore that our methodology can be extended to other tasks just by adapting the search space accordingly, while the search strategy and overall training procedure remain unchanged.

By using large assorted operator sets on time series classification datasets, POPNASv3 has been able to find innovative architectures, with mixed usage of convolutions, dilated convolutions, and recurrent neural network units.
The experiments provide evidence that POPNASv3 can determine the most suitable operators to process each dataset, sampling a minimal amount of architectures to converge to the optimal search space regions, making it feasible to explore large search spaces in a short amount of time.
The experiments show that POPNASv3 is able to determine the most suitable operators to process each dataset, converging to the optimal regions of the search space by sampling a small number of architectures, making feasible the exploration of large search spaces in a short period of time.

\section*{Acknowledgments}
The European Commission has partially funded this work under the H2020 grant N. 101016577 AI-SPRINT: AI in Secure Privacy-pReserving computINg conTinuum.

\bibliographystyle{unsrt}  
\bibliography{bibliography}

\begin{thebibliography}{10}

\bibitem{goodfellow2016deep}
Ian Goodfellow, Yoshua Bengio, and Aaron Courville.
\newblock {\em Deep learning}.
\newblock 2016.

\bibitem{Alzubaidi2021}
Laith Alzubaidi, Jinglan Zhang, Amjad~J. Humaidi, Ayad Al-Dujaili, Ye~Duan,
  Omran Al-Shamma, J.~Santamar{\'i}a, Mohammed~A. Fadhel, Muthana Al-Amidie,
  and Laith Farhan.
\newblock Review of deep learning: concepts, cnn architectures, challenges,
  applications, future directions.
\newblock {\em Journal of Big Data}, 8(1):53, Mar 2021.

\bibitem{Dargan2020}
Shaveta Dargan, Munish Kumar, Maruthi~Rohit Ayyagari, and Gulshan Kumar.
\newblock A survey of deep learning and its applications: A new paradigm to
  machine learning.
\newblock {\em Archives of Computational Methods in Engineering},
  27(4):1071--1092, Sep 2020.

\bibitem{AlexNet}
Alex Krizhevsky, Ilya Sutskever, and Geoffrey~E. Hinton.
\newblock Imagenet classification with deep convolutional neural networks.
\newblock In {\em Proceedings of the 25th International Conference on Neural
  Information Processing Systems - Volume 1}, NIPS'12, page 1097–1105, Red
  Hook, NY, USA, 2012. Curran Associates Inc.

\bibitem{VGG}
Karen Simonyan and Andrew Zisserman.
\newblock Very deep convolutional networks for large-scale image recognition,
  2014.

\bibitem{ResNet}
Kaiming He, Xiangyu Zhang, Shaoqing Ren, and Jian Sun.
\newblock Deep residual learning for image recognition, 2015.

\bibitem{DBLP:journals/corr/abs-1904-12054}
Marc{-}Andr{\'{e}} Z{\"{o}}ller and Marco~F. Huber.
\newblock Survey on automated machine learning.
\newblock {\em CoRR}, abs/1904.12054, 2019.

\bibitem{zoph2017neural}
Barret Zoph and Quoc~V. Le.
\newblock Neural architecture search with reinforcement learning, 2017.

\bibitem{OneShotUniform}
Zichao Guo, Xiangyu Zhang, Haoyuan Mu, Wen Heng, Zechun Liu, Yichen Wei, and
  Jian Sun.
\newblock Single path one-shot neural architecture search with uniform
  sampling, 2019.

\bibitem{UnderstandingOneShot}
Gabriel Bender, Pieter-Jan Kindermans, Barret Zoph, Vijay Vasudevan, and Quoc
  Le.
\newblock Understanding and simplifying one-shot architecture search.
\newblock In Jennifer Dy and Andreas Krause, editors, {\em Proceedings of the
  35th International Conference on Machine Learning}, volume~80 of {\em
  Proceedings of Machine Learning Research}, pages 550--559. PMLR, 10--15 Jul
  2018.

\bibitem{smbo}
Frank Hutter, Holger~H. Hoos, and Kevin Leyton-Brown.
\newblock Sequential model-based optimization for general algorithm
  configuration.
\newblock In Carlos A.~Coello Coello, editor, {\em Learning and Intelligent
  Optimization}, pages 507--523, Berlin, Heidelberg, 2011. Springer Berlin
  Heidelberg.

\bibitem{elsken2019neural}
Thomas Elsken, Jan~Hendrik Metzen, and Frank Hutter.
\newblock Neural architecture search: A survey.
\newblock {\em The Journal of Machine Learning Research}, 20(1):1997--2017,
  2019.

\bibitem{real2019regularized}
Esteban Real, Alok Aggarwal, Yanping Huang, and Quoc~V Le.
\newblock Regularized evolution for image classifier architecture search, 2019.

\bibitem{zoph2018learning}
Barret Zoph, Vijay Vasudevan, Jonathon Shlens, and Quoc~V. Le.
\newblock Learning transferable architectures for scalable image recognition,
  2018.

\bibitem{Inception}
Christian Szegedy, Wei Liu, Yangqing Jia, Pierre Sermanet, Scott Reed, Dragomir
  Anguelov, Dumitru Erhan, Vincent Vanhoucke, and Andrew Rabinovich.
\newblock Going deeper with convolutions, 2014.

\bibitem{liu2018progressive}
Chenxi Liu, Barret Zoph, Maxim Neumann, Jonathon Shlens, Wei Hua, Li-Jia Li,
  Li~Fei-Fei, Alan Yuille, Jonathan Huang, and Kevin Murphy.
\newblock Progressive neural architecture search, 2018.

\bibitem{NSGANet}
Zhichao Lu, Ian Whalen, Vishnu Boddeti, Yashesh Dhebar, Kalyanmoy Deb, Erik
  Goodman, and Wolfgang Banzhaf.
\newblock Nsga-net: Neural architecture search using multi-objective genetic
  algorithm, 2018.

\bibitem{ENAS}
Hieu Pham, Melody~Y. Guan, Barret Zoph, Quoc~V. Le, and Jeff Dean.
\newblock Efficient neural architecture search via parameter sharing, 2018.

\bibitem{elsken2018efficient}
Thomas Elsken, Jan~Hendrik Metzen, and Frank Hutter.
\newblock Efficient multi-objective neural architecture search via lamarckian
  evolution.
\newblock {\em arXiv preprint arXiv:1804.09081}, 2018.

\bibitem{hsu2018monas}
Chi-Hung Hsu, Shu-Huan Chang, Jhao-Hong Liang, Hsin-Ping Chou, Chun-Hao Liu,
  Shih-Chieh Chang, Jia-Yu Pan, Yu-Ting Chen, Wei Wei, and Da-Cheng Juan.
\newblock Monas: Multi-objective neural architecture search using reinforcement
  learning.
\newblock {\em arXiv preprint arXiv:1806.10332}, 2018.

\bibitem{POPNAS}
Eugenio Lomurno, Stefano Samele, Matteo Matteucci, and Danilo Ardagna.
\newblock {\em Pareto-Optimal Progressive Neural Architecture Search}, page
  1726–1734.
\newblock Association for Computing Machinery, New York, NY, USA, 2021.

\bibitem{POPNASv2}
Andrea Falanti, Eugenio Lomurno, Stefano Samele, Danilo Ardagna, and Matteo
  Matteucci.
\newblock Popnasv2: An efficient multi-objective neural architecture search
  technique.
\newblock In {\em 2022 International Joint Conference on Neural Networks
  (IJCNN)}, pages 1--8, 2022.

\bibitem{liu2019darts}
Hanxiao Liu, Karen Simonyan, and Yiming Yang.
\newblock Darts: Differentiable architecture search, 2019.

\bibitem{cai2020OFA}
Han Cai, Chuang Gan, Tianzhe Wang, Zhekai Zhang, and Song Han.
\newblock Once for all: Train one network and specialize it for efficient
  deployment.
\newblock In {\em International Conference on Learning Representations}, 2020.

\bibitem{UCRArchive2018}
Hoang~Anh Dau, Eamonn Keogh, Kaveh Kamgar, Chin-Chia~Michael Yeh, Yan Zhu,
  Shaghayegh Gharghabi, Chotirat~Ann Ratanamahatana, Yanping, Bing Hu, Nurjahan
  Begum, Anthony Bagnall, Abdullah Mueen, Gustavo Batista, and Hexagon-ML.
\newblock The ucr time series classification archive, October 2018.
\newblock \url{https://www.cs.ucr.edu/~eamonn/time_series_data_2018/}.

\bibitem{UEAArchive2018}
Anthony Bagnall, Hoang~Anh Dau, Jason Lines, Michael Flynn, James Large, Aaron
  Bostrom, Paul Southam, and Eamonn Keogh.
\newblock The uea multivariate time series classification archive, 2018, 2018.

\bibitem{InceptionTime}
Hassan Ismail~Fawaz, Benjamin Lucas, Germain Forestier, Charlotte Pelletier,
  Daniel~F. Schmidt, Jonathan Weber, Geoffrey~I. Webb, Lhassane Idoumghar,
  Pierre-Alain Muller, and Fran{\c{c}}ois Petitjean.
\newblock Inceptiontime: Finding alexnet for time series classification.
\newblock {\em Data Mining and Knowledge Discovery}, 34(6):1936--1962, Nov
  2020.

\bibitem{ROCKET}
Angus Dempster, Fran{\c{c}}ois Petitjean, and Geoffrey~I. Webb.
\newblock Rocket: exceptionally fast and accurate time series classification
  using random convolutional kernels.
\newblock {\em Data Mining and Knowledge Discovery}, 34(5):1454--1495, Sep
  2020.

\bibitem{HIVE-COTEv2}
Matthew Middlehurst, James Large, Michael Flynn, Jason Lines, Aaron Bostrom,
  and Anthony Bagnall.
\newblock Hive-cote 2.0: a new meta ensemble for time series classification.
\newblock {\em Machine Learning}, 110(11):3211--3243, Dec 2021.

\bibitem{NAS-T}
Hojjat Rakhshani, Hassan Ismail~Fawaz, Lhassane Idoumghar, Germain Forestier,
  Julien Lepagnot, Jonathan Weber, Mathieu Brévilliers, and Pierre-Alain
  Muller.
\newblock Neural architecture search for time series classification.
\newblock In {\em 2020 International Joint Conference on Neural Networks
  (IJCNN)}, pages 1--8, 2020.

\bibitem{ResNet-for-TSC}
Zhiguang Wang, Weizhong Yan, and Tim Oates.
\newblock Time series classification from scratch with deep neural networks: A
  strong baseline, 2016.

\bibitem{hochreiter1997long}
Sepp Hochreiter and J{\"u}rgen Schmidhuber.
\newblock Long short-term memory.
\newblock {\em Neural computation}, 9(8):1735--1780, 1997.

\bibitem{GRU}
Kyunghyun Cho, Bart van Merrienboer, Caglar Gulcehre, Dzmitry Bahdanau, Fethi
  Bougares, Holger Schwenk, and Yoshua Bengio.
\newblock Learning phrase representations using rnn encoder-decoder for
  statistical machine translation, 2014.

\bibitem{BatchNorm}
Sergey Ioffe and Christian Szegedy.
\newblock Batch normalization: Accelerating deep network training by reducing
  internal covariate shift, 2015.

\bibitem{ramachandran2017searching}
Prajit Ramachandran, Barret Zoph, and Quoc~V. Le.
\newblock Searching for activation functions, 2017.

\bibitem{loshchilov2019decoupled}
Ilya Loshchilov and Frank Hutter.
\newblock Decoupled weight decay regularization, 2019.

\bibitem{loshchilov2016sgdr}
Ilya Loshchilov and Frank Hutter.
\newblock Sgdr: Stochastic gradient descent with warm restarts.
\newblock {\em arXiv preprint arXiv:1608.03983}, 2016.

\bibitem{SelfAttention}
Jianpeng Cheng, Li~Dong, and Mirella Lapata.
\newblock Long short-term memory-networks for machine reading, 2016.

\bibitem{prokhorenkova2019catboost}
Liudmila Prokhorenkova, Gleb Gusev, Aleksandr Vorobev, Anna~Veronika Dorogush,
  and Andrey Gulin.
\newblock Catboost: unbiased boosting with categorical features, 2019.

\bibitem{SHAP}
Scott~M Lundberg and Su-In Lee.
\newblock A unified approach to interpreting model predictions.
\newblock In I.~Guyon, U.~V. Luxburg, S.~Bengio, H.~Wallach, R.~Fergus,
  S.~Vishwanathan, and R.~Garnett, editors, {\em Advances in Neural Information
  Processing Systems 30}, pages 4765--4774. Curran Associates, Inc., 2017.

\bibitem{fractalnet}
Gustav Larsson, Michael Maire, and Gregory Shakhnarovich.
\newblock Fractalnet: Ultra-deep neural networks without residuals, 2016.

\bibitem{Cutout}
Terrance DeVries and Graham~W. Taylor.
\newblock Improved regularization of convolutional neural networks with cutout,
  2017.

\bibitem{helber2019eurosat}
Patrick Helber, Benjamin Bischke, Andreas Dengel, and Damian Borth.
\newblock Eurosat: A novel dataset and deep learning benchmark for land use and
  land cover classification, 2019.

\bibitem{Krizhevsky09learningmultiple}
Alex Krizhevsky.
\newblock Learning multiple layers of features from tiny images.
\newblock Technical report, University of Toronto, 2009.

\bibitem{xiao2017fashionmnist}
Han Xiao, Kashif Rasul, and Roland Vollgraf.
\newblock Fashion-mnist: a novel image dataset for benchmarking machine
  learning algorithms, 2017.

\bibitem{tensorflow2015-whitepaper}
Mart\'{i}n Abadi, Ashish Agarwal, Paul Barham, Eugene Brevdo, Zhifeng Chen,
  Craig Citro, Greg~S. Corrado, Andy Davis, Jeffrey Dean, Matthieu Devin,
  Sanjay Ghemawat, Ian Goodfellow, Andrew Harp, Geoffrey Irving, Michael Isard,
  Yangqing Jia, Rafal Jozefowicz, Lukasz Kaiser, Manjunath Kudlur, Josh
  Levenberg, Dandelion Man\'{e}, Rajat Monga, Sherry Moore, Derek Murray, Chris
  Olah, Mike Schuster, Jonathon Shlens, Benoit Steiner, Ilya Sutskever, Kunal
  Talwar, Paul Tucker, Vincent Vanhoucke, Vijay Vasudevan, Fernanda Vi\'{e}gas,
  Oriol Vinyals, Pete Warden, Martin Wattenberg, Martin Wicke, Yuan Yu, and
  Xiaoqiang Zheng.
\newblock {TensorFlow}: Large-scale machine learning on heterogeneous systems,
  2015.
\newblock Software available from tensorflow.org.

\bibitem{LabelSmoothingHelp}
Rafael Müller, Simon Kornblith, and Geoffrey Hinton.
\newblock When does label smoothing help?, 2019.

\bibitem{kingma2017adam}
Diederik~P. Kingma and Jimmy Ba.
\newblock Adam: A method for stochastic optimization, 2017.

\bibitem{CRNN-music-class}
Keunwoo Choi, György Fazekas, Mark Sandler, and Kyunghyun Cho.
\newblock Convolutional recurrent neural networks for music classification.
\newblock In {\em 2017 IEEE International Conference on Acoustics, Speech and
  Signal Processing (ICASSP)}, pages 2392--2396, 2017.

\bibitem{CRNN-ECG-class}
Martin Zihlmann, Dmytro Perekrestenko, and Michael Tschannen.
\newblock Convolutional recurrent neural networks for electrocardiogram
  classification.
\newblock In {\em 2017 Computing in Cardiology (CinC)}, pages 1--4, 2017.

\bibitem{JMLR:v7:demsar06a}
Janez Dem{\v{s}}ar.
\newblock Statistical comparisons of classifiers over multiple data sets.
\newblock {\em Journal of Machine Learning Research}, 7(1):1--30, 2006.

\bibitem{JMLR:v9:garcia08a}
Salvador Garc{{\'i}}a and Francisco Herrera.
\newblock An extension on ``statistical comparisons of classifiers over
  multiple data sets'' for all pairwise comparisons.
\newblock {\em Journal of Machine Learning Research}, 9(89):2677--2694, 2008.

\bibitem{JMLR:v17:benavoli16a}
Alessio Benavoli, Giorgio Corani, and Francesca Mangili.
\newblock Should we really use post-hoc tests based on mean-ranks?
\newblock {\em Journal of Machine Learning Research}, 17(5):1--10, 2016.

\bibitem{IsmailFawaz2018deep}
Hassan Ismail~Fawaz, Germain Forestier, Jonathan Weber, Lhassane Idoumghar, and
  Pierre-Alain Muller.
\newblock Deep learning for time series classification: a review.
\newblock {\em Data Mining and Knowledge Discovery}, 33(4):917--963, 2019.

\end{thebibliography}

\appendix
\section{Appendix}
\subsection{POPNASv3 Top Cells}
Here we report an illustration of all the top-1 cell structures found by POPNASv3 on the datasets considered in the experiments section.

\begin{figure}[ht]
    \subfloat[Ford A]{
	    \label{FIG:cell-ford-a}
	    \includegraphics[width=0.55\textwidth]{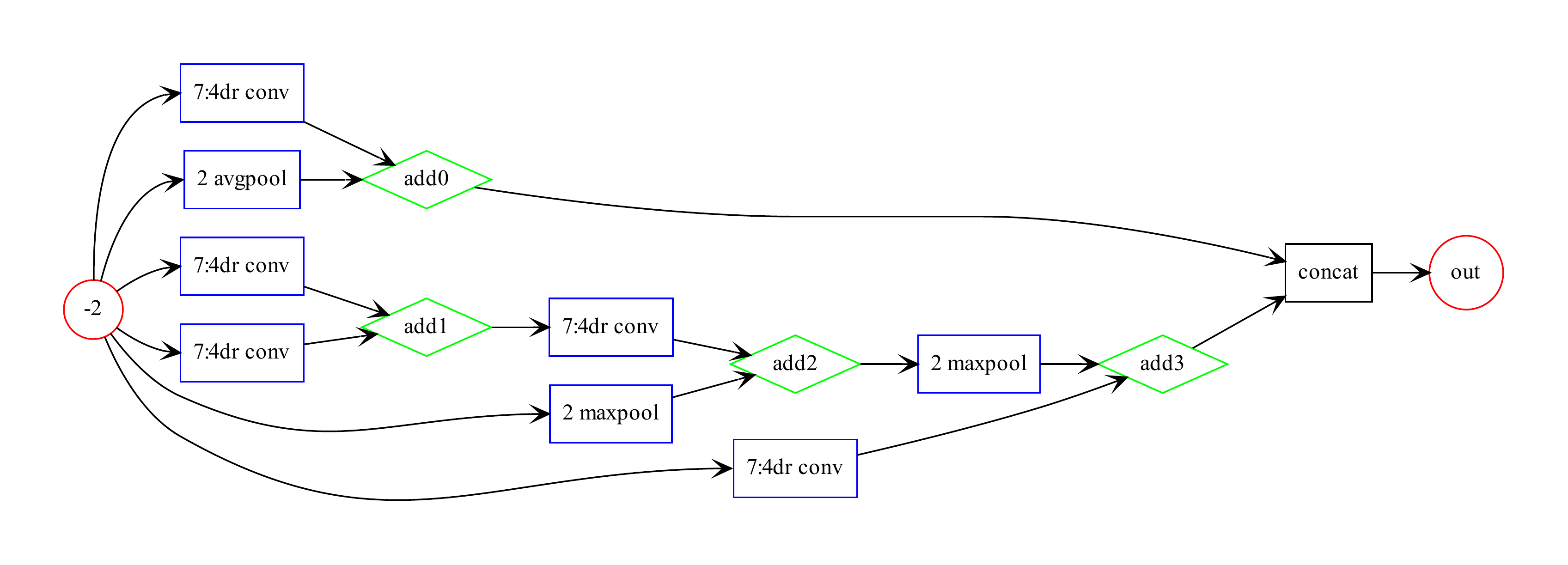}
	}
    \hfill
	\subfloat[ElectricDevices]{
	    \label{FIG:cell-ed}
	    \includegraphics[width=0.41\textwidth]{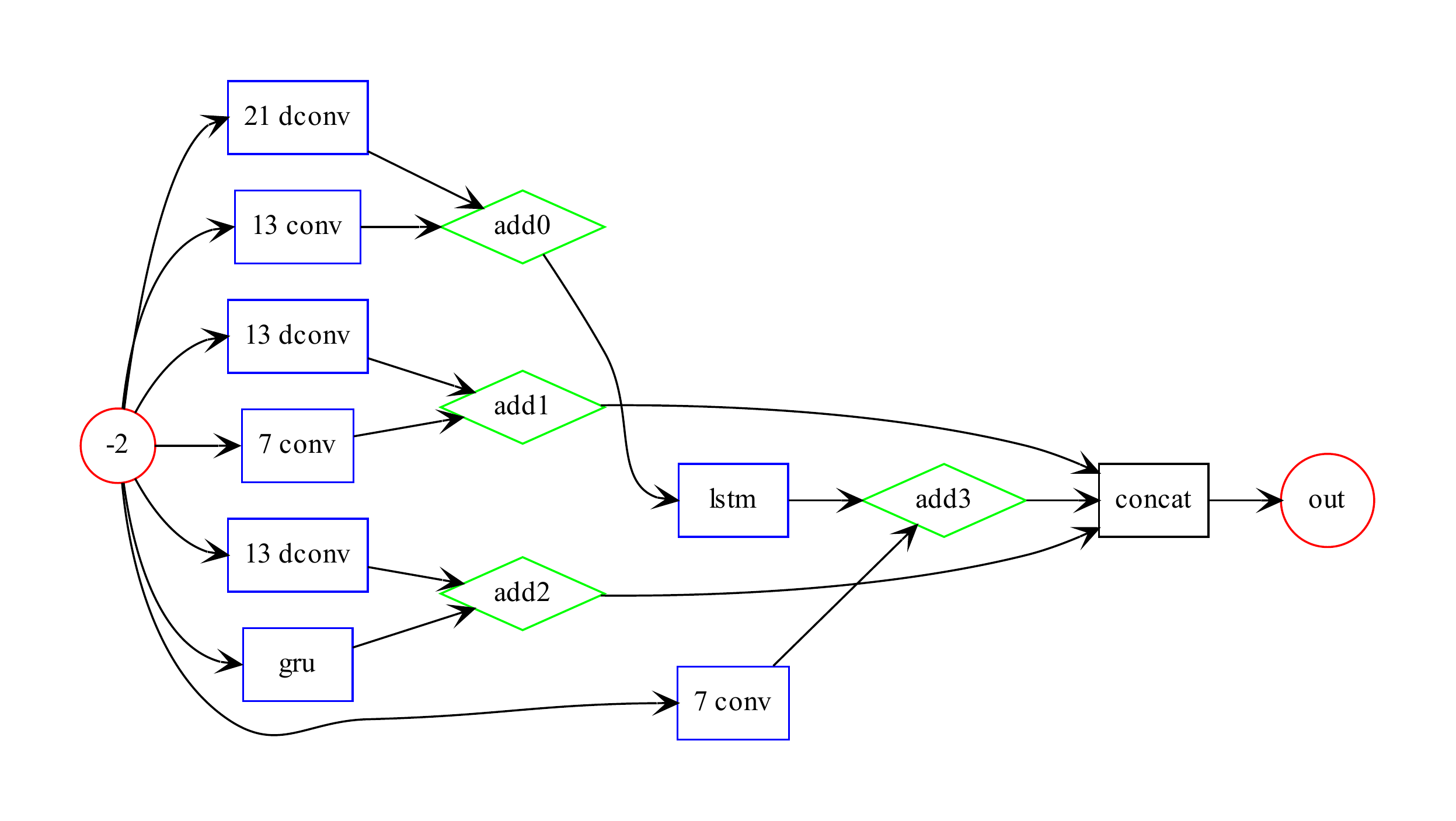}
	}
    \hfill
    \subfloat[FaceDetection]{
	    \label{FIG:cell-face-detection}
	    \includegraphics[width=0.52\textwidth]{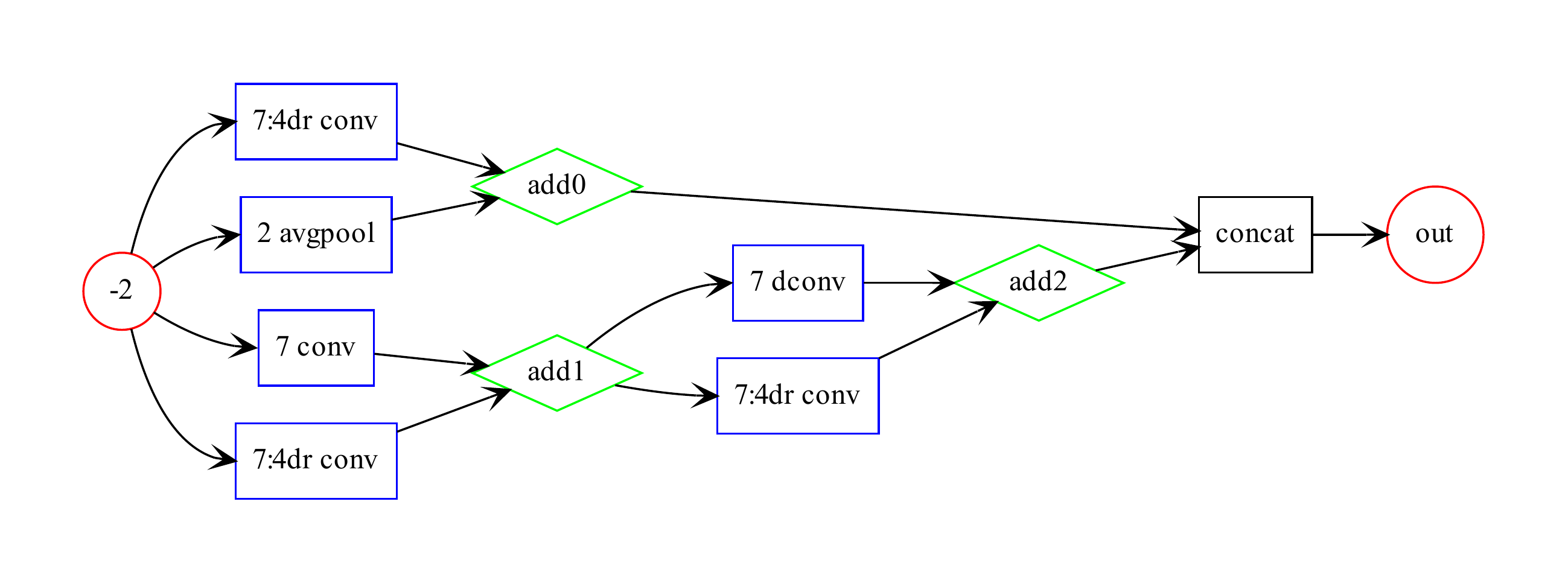}
	} 
    \hfill
	\subfloat[LSST]{
	    \label{FIG:cell-lsst}
	    \includegraphics[width=0.44\textwidth]{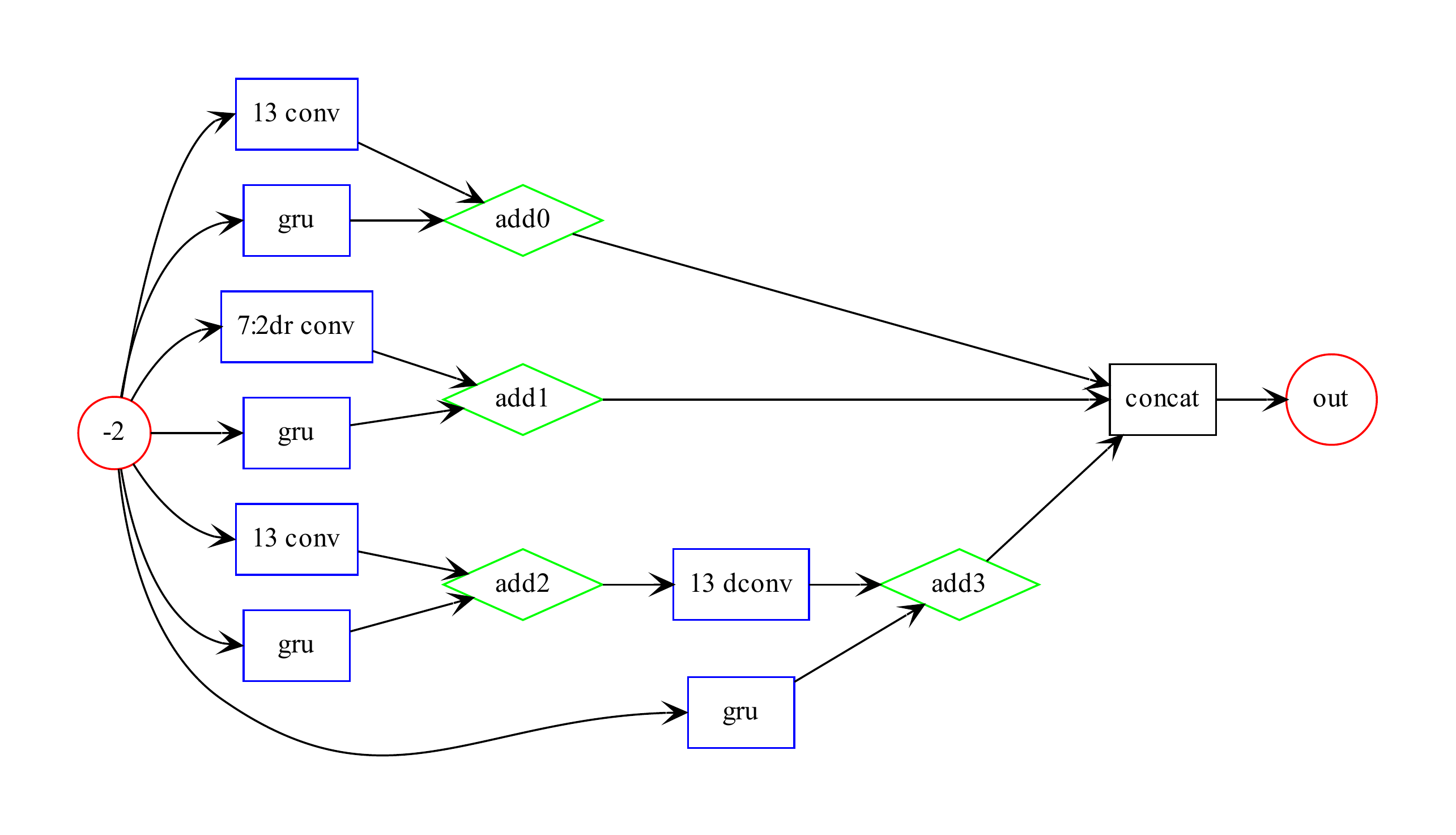}
	}
	\caption{Cell motifs used in the best architectures found on time series classification datasets.}
	\label{FIG:TSC-best-cells}
\end{figure}

\begin{figure}[ht]
	\centering
	\subfloat[CIFAR10]{
	    \label{FIG:cell-c10}
	    \includegraphics[width=0.60\textwidth]{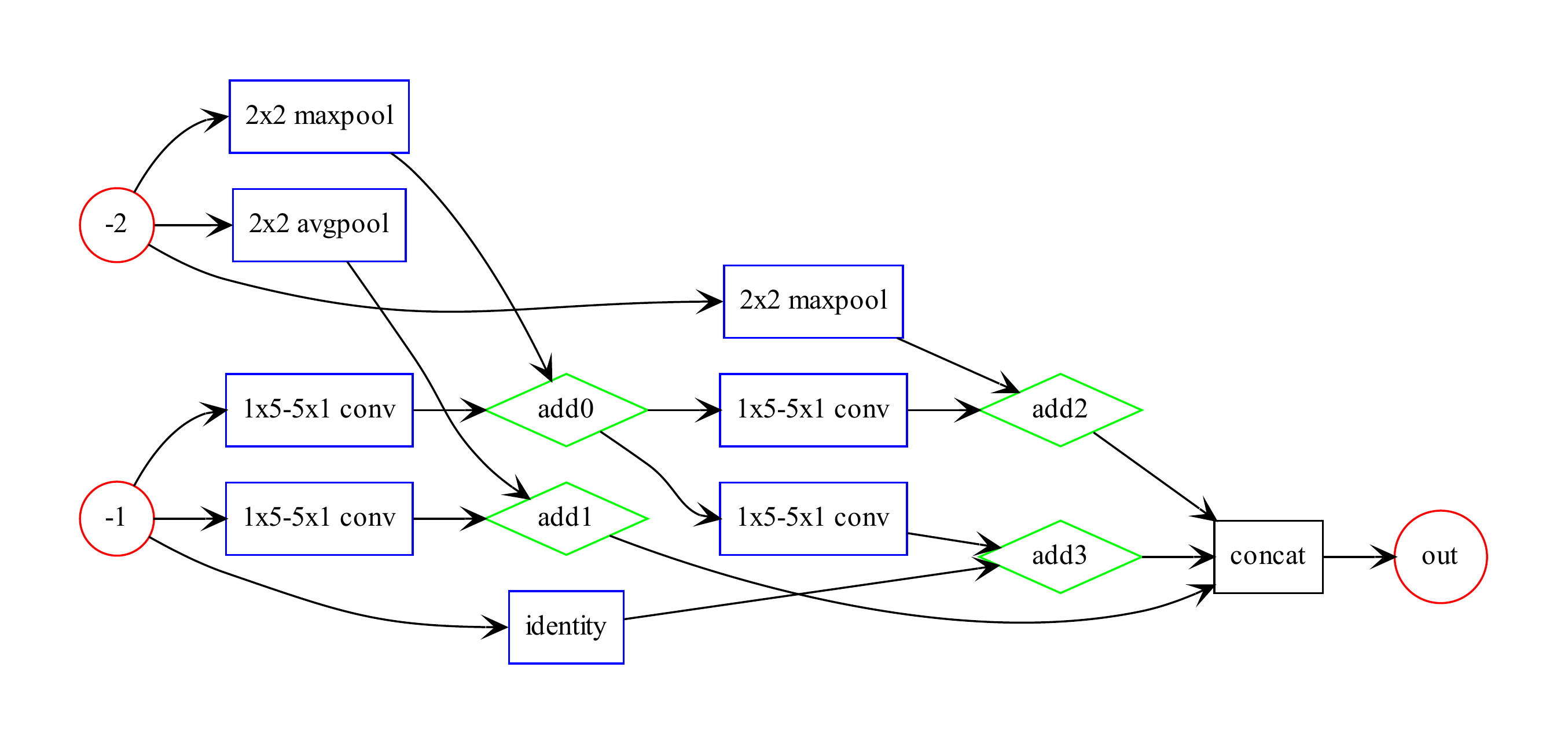}
	}
    \hfill
	\subfloat[CIFAR100]{
	    \label{FIG:cell-c100}
	    \includegraphics[width=0.35\textwidth]{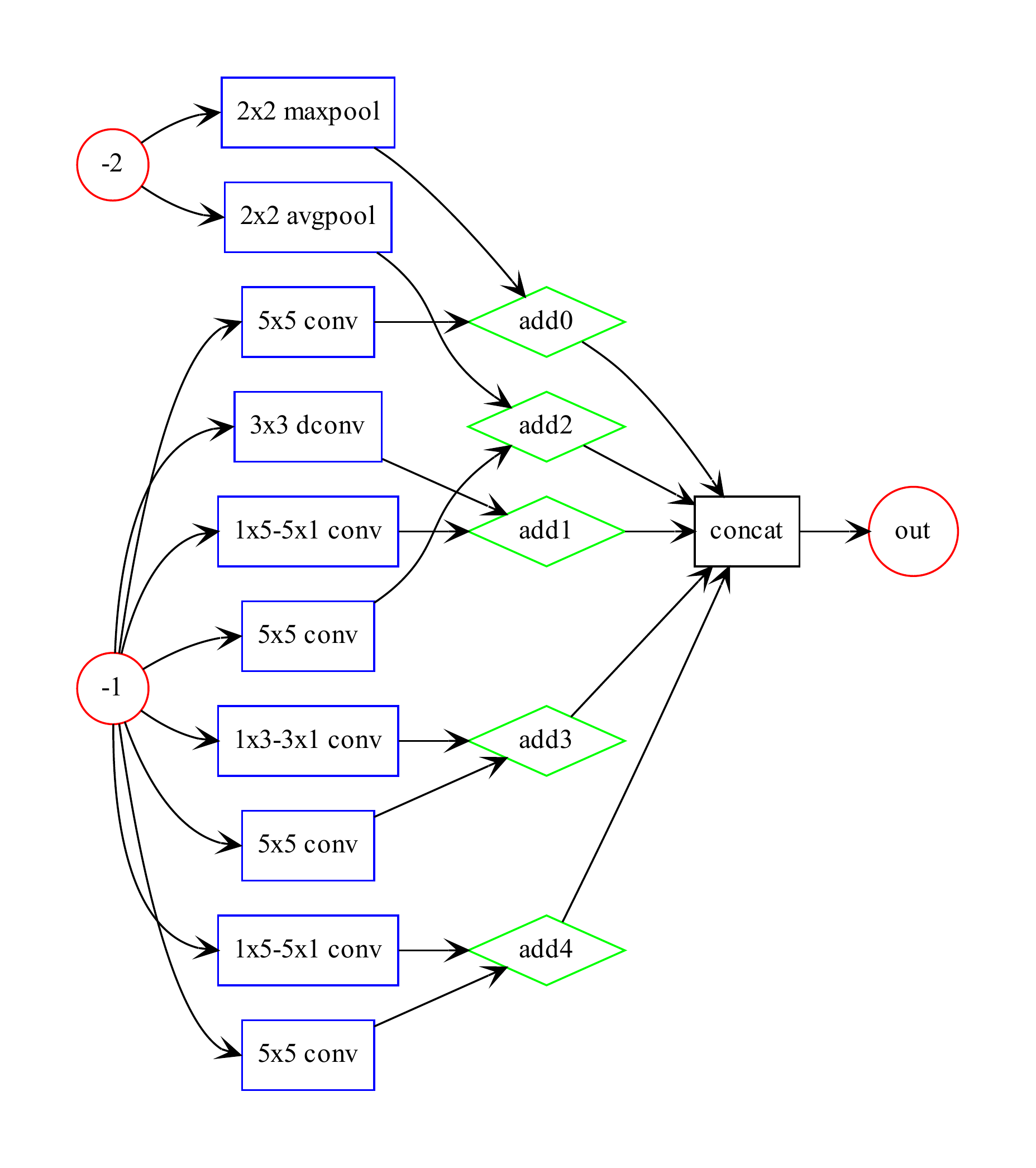}
	}
    \hfill
	\subfloat[Fashion MNIST]{
	    \label{FIG:cell-fmnist}
	    \includegraphics[width=0.60\textwidth]{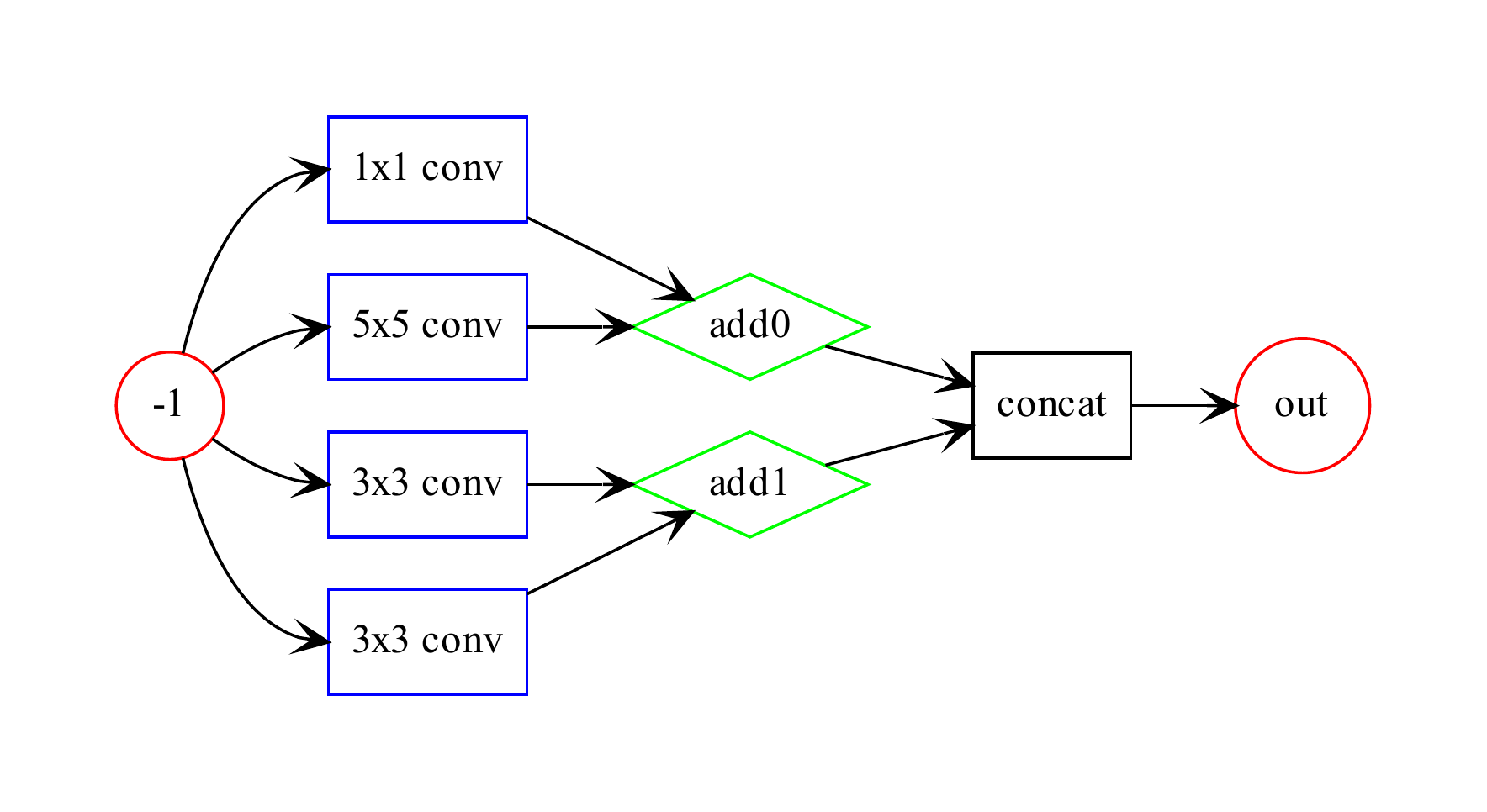}
	}
    \hfill
    \subfloat[EuroSAT]{
	    \label{FIG:cell-eurosat}
	    \includegraphics[width=0.35\textwidth]{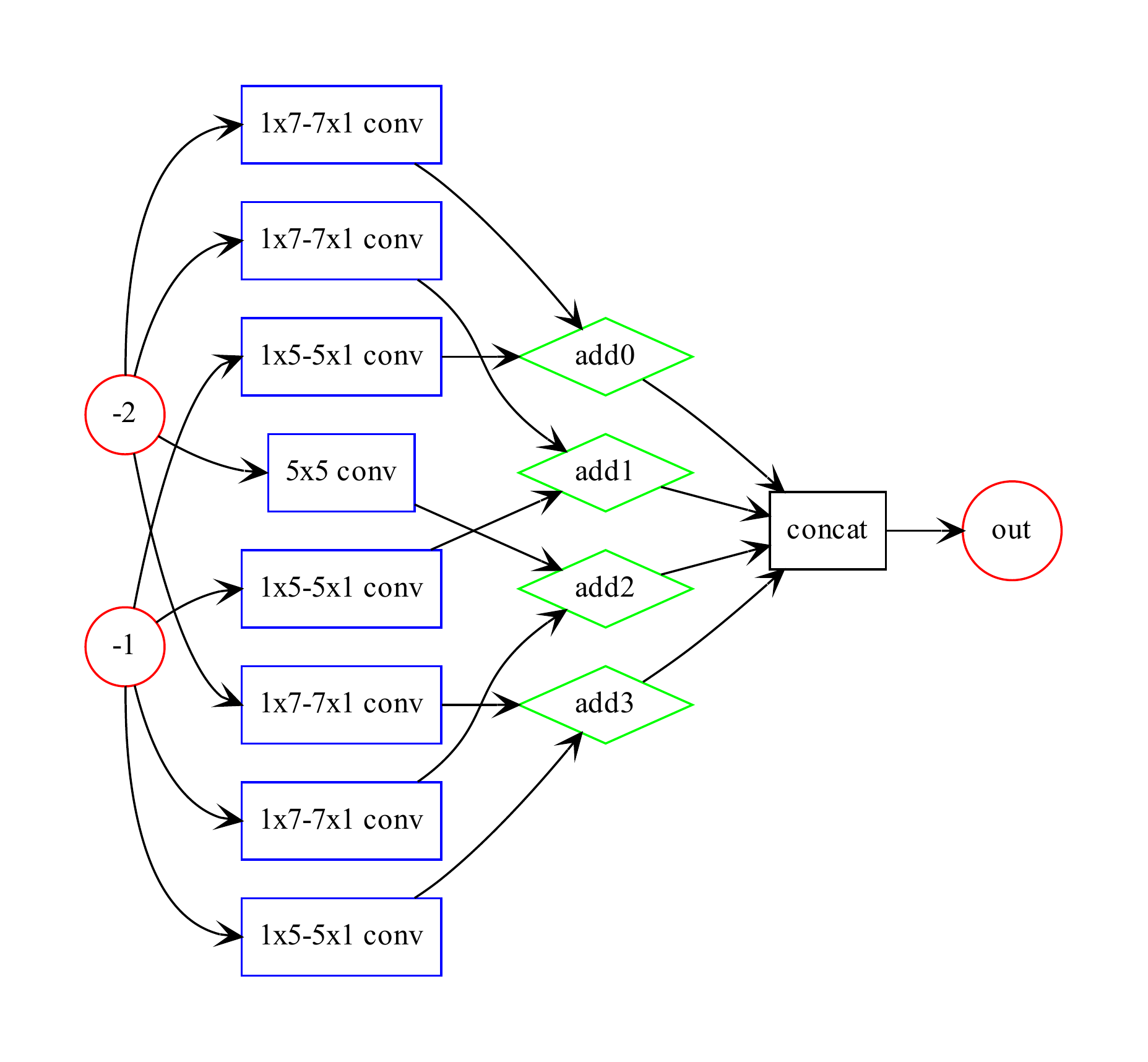}
	} 
	\caption{Cell motifs used in the best architectures found on image classification datasets.}
	\label{FIG:IC-best-cells}
\end{figure}

\clearpage
\subsection{Extended Results on UCR/UEA archives} \label{SEC:TSC-extended-apx}%

This section presents the results of the experiments performed on the extended selection of time series classification datasets, composed of 45 univariate datasets and 4 multivariate ones.
These datasets contain at least 360 samples in the training set, and their samples have equal lengths.
The results have been analyzed in Section~\ref{SEC:TSC-results}, here we just report a summary of the results under multiple tables.
Multivariate datasets are separated from the univariate datasets and presented at the bottom of each table.

\begin{table}[h]
\caption{Search results and top validation accuracies found on the extended UCR / UEA datasets selection.}
\label{TAB:ext-TSC-search}
\centering
\footnotesize
\begin{tabularx}{\linewidth}{@{} l YYYl @{}} 
\toprule
Dataset & \# Networks & Top Accuracy & Top-5 Cells Accuracy & Search Time \\
\midrule
Adiac & 417 & 0.556 & 0.530 & 3h 10m \\
ChlorineConcentration & 404 & 0.639 & 0.630 & 3h 23m \\
CricketX & 478 & 0.825 & 0.816 & 4h 18m \\
CricketY & 501 & 0.857 & 0.838 & 5h 1m \\
CricketZ & 445 & 0.778 & 0.752 & 3h 46m \\
Crop & 496 & 0.775 & 0.772 & 8h 9m \\
DistalPhalanxOutlineAgeGroup & 428 & 0.900 & 0.883 & 3h 28m \\
DistalPhalanxOutlineCorrect & 420 & 0.950 & 0.937 & 3h 12m \\
DistalPhalanxTW & 417 & 0.850 & 0.840 & 3h 14m \\
ECG5000 & 432 & 0.983 & 0.970 & 3h 59m \\
EOGHorizontalSignal & 509 & 0.919 & 0.877 & 13h 28m \\
EOGVerticalSignal & 525 & 0.677 & 0.665 & 14h 21m \\
ElectricDevices & 507 & 0.920 & 0.918 & 13h 55m \\
EthanolLevel & 446 & 0.525 & 0.502 & 6h 37m \\
FaceAll & 447 & 1.000 & 1.000 & 3h 35m \\
FiftyWords & 451 & 0.844 & 0.819 & 4h 13m \\
FordA & 475 & 0.970 & 0.966 & 7h 38m \\
FordB & 467 & 0.967 & 0.965 & 7h 19m \\
HandOutlines & 482 & 0.920 & 0.904 & 13h 29m \\
LargeKitchenAppliances & 487 & 1.000 & 0.987 & 7h 46m \\
MedicalImages & 477 & 0.787 & 0.780 & 4h 18m \\
MiddlePhalanxOutlineAgeGroup & 413 & 0.833 & 0.827 & 3h 9m \\
MiddlePhalanxOutlineCorrect & 425 & 0.867 & 0.850 & 3h 31m \\
MiddlePhalanxTW & 405 & 0.734 & 0.728 & 3h 8m \\
NonInvasiveFetalECGThorax1 & 504 & 0.811 & 0.759 & 6h 35m \\
NonInvasiveFetalECGThorax2 & 456 & 0.861 & 0.853 & 5h 48m \\
PhalangesOutlinesCorrect & 426 & 0.878 & 0.859 & 3h 28m \\
ProximalPhalanxOutlineAgeGroup & 435 & 0.933 & 0.933 & 3h 37m \\
ProximalPhalanxOutlineCorrect & 430 & 0.933 & 0.923 & 3h 19m \\
ProximalPhalanxTW & 473 & 0.867 & 0.853 & 4h 26m \\
RefrigerationDevices & 438 & 0.783 & 0.763 & 5h 52m \\
ScreenType & 428 & 0.583 & 0.567 & 5h 41m \\
SemgHandMovementCh2 & 509 & 0.656 & 0.641 & 11h 9m \\
SemgHandSubjectCh2 & 516 & 0.859 & 0.850 & 8h 36m \\
ShapesAll & 509 & 0.883 & 0.857 & 6h 44m \\
SmallKitchenAppliances & 446 & 0.850 & 0.830 & 6h 25m \\
StarLightCurves & 511 & 0.980 & 0.980 & 9h 1m \\
Strawberry & 452 & 0.952 & 0.948 & 4h 19m \\
SwedishLeaf & 475 & 0.933 & 0.917 & 4h 27m \\
TwoPatterns & 484 & 1.000 & 1.000 & 4h 42m \\
UWaveGestureLibraryAll & 567 & 0.978 & 0.951 & 9h 4m \\
UWaveGestureLibraryX & 491 & 0.844 & 0.827 & 4h 40m \\
UWaveGestureLibraryY & 488 & 0.800 & 0.793 & 4h 38m \\
UWaveGestureLibraryZ & 421 & 0.789 & 0.782 & 3h 43m \\
Wafer & 465 & 1.000 & 1.000 & 4h 7m \\
\midrule
FaceDetection & 467 & 0.800 & 0.794 & 7h 25m \\
LSST & 471 & 0.663 & 0.647 & 5h 15m \\
PenDigits & 469 & 0.999 & 0.999 & 7h 56m \\
PhonemeSpectra & 572 & 0.358 & 0.351 & 11h 9m \\
\bottomrule
\end{tabularx}
\end{table}

\begin{table}[t]
\caption{Final training results on the extended UCR / UEA datasets selection. The macro-architecture parameters and the cell specifications are listed together with the accuracy reached on the default test split.}
\label{TAB:ext-TSC-final}
\centering
\scriptsize
\begin{tabularx}{\linewidth}{@{} p{3.5cm} ccccYY p{0.50\linewidth} @{}} 
\toprule
Dataset & B & M & N & F & Params & Acc. & Cell specification \\
 & & & & & (M) & & \\
\midrule
Adiac & 1 & 3 & 2 & 42 & 2.47 & 0.821 & {[}(-2, 'gru', -1, '21 conv')] \\
ChlorineConcentration & 1 & 3 & 4 & 48 & 1.56 & 0.810 & {[}(-2, '13 dconv', -2, '13 conv')] \\
CricketX & 4 & 3 & 2 & 48 & 2.17 & 0.821 & {[}(-2, '21 dconv', -2, '7:4dr conv');(-2, '21 dconv', -2, '7:4dr conv');(-2, '13 conv', -2, '2 avgpool');(2, '13 dconv', 2, '2 avgpool')] \\
CricketY & 3 & 4 & 2 & 24 & 2.30 & 0.862 & {[}(-2, 'identity', -2, '7:4dr conv');(-2, '2 maxpool', 0, '7:4dr conv');(0, 'gru', 1, '7:4dr conv')] \\
CricketZ & 1 & 3 & 2 & 24 & 0.38 & 0.831 & {[}(-2, '7:4dr conv', -1, '7:4dr conv')] \\
Crop & 1 & 3 & 2 & 24 & 0.25 & 0.782 & {[}(-2, '21 dconv', -1, '7:4dr conv')] \\
DistalPhalanxOutlineAgeGroup & 1 & 3 & 2 & 24 & 0.14 & 0.691 & {[}(-2, '13 dconv', -2, 'gru')] \\
DistalPhalanxOutlineCorrect & 3 & 4 & 4 & 48 & 2.78 & 0.793 & {[}(-2, 'identity', -1, '2 maxpool');(-2, '2 maxpool', -2, '2 maxpool');(1, 'identity', 1, '2 maxpool')] \\
DistalPhalanxTW & 1 & 3 & 4 & 48 & 1.56 & 0.712 & {[}(-1, '7:4dr conv', -1, '2 avgpool')] \\
ECG5000 & 1 & 3 & 2 & 24 & 0.20 & 0.927 & {[}(-2, 'gru', -1, '2 avgpool')] \\
EOGHorizontalSignal & 5 & 3 & 3 & 42 & 2.89 & 0.655 & {[}(-2, '2 maxpool', -2, '2 maxpool');(0, '2 maxpool', 0, '2 avgpool');(0, '7:4dr conv', 1, '7:4dr conv');(2, '21 dconv', 2, '2 maxpool');(-2, '21 conv', -2, 'gru')] \\
EOGVerticalSignal & 1 & 3 & 2 & 48 & 1.65 & 0.591 & {[}(-1, '7:4dr conv', -1, 'gru')] \\
ElectricDevices & 4 & 3 & 2 & 24 & 2.15 & 0.744 & {[}(-1, '7 conv', -1, '13 conv');(-1, '13 conv', -1, '13 conv');(-1, '7 conv', -1, 'gru');(1, '7 dconv', 1, 'gru')] \\
EthanolLevel & 1 & 4 & 2 & 20 & 2.31 & 0.850 & {[}(-2, 'gru', -1, '21 conv')] \\
FaceAll & 1 & 3 & 2 & 24 & 0.08 & 0.956 & {[}(-2, '7 dconv', -1, '21 dconv')] \\
FiftyWords & 3 & 4 & 2 & 24 & 1.38 & 0.862 & {[}(-2, 'identity', -2, '2 avgpool');(-2, 'identity', -2, '2 avgpool');(0, '7:4dr conv', 1, '7:4dr conv')] \\
FordA & 2 & 4 & 2 & 36 & 2.58 & 0.953 & {[}(-2, 'identity', -2, '7:4dr conv');(-2, 'identity', -2, '7:4dr conv')] \\
FordB & 1 & 4 & 3 & 24 & 2.34 & 0.843 & {[}(-1, '7:4dr conv', -1, 'gru')] \\
HandOutlines & 4 & 4 & 2 & 48 & 2.63 & 0.927 & 1{[}(-2, 'identity', -2, '2 avgpool');(-2, 'identity', -2, '2 avgpool');(0, '2 avgpool', 1, 'gru');(0, 'identity', 2, 'identity')] \\
LargeKitchenAppliances & 5 & 3 & 2 & 48 & 3.01 & 0.904 & {[}(-2, 'identity', -2, 'identity');(-2, 'identity', -2, 'identity');(-2, '7 conv', 1, '7:4dr conv');(-2, '21 conv', 1, '13 dconv');(-2, '21 dconv', 1, '13 dconv')] \\
MedicalImages & 3 & 3 & 2 & 24 & 0.82 & 0.784 & {[}(-2, '7:4dr conv', -2, '7:4dr conv');(-2, '7 dconv', 0, 'identity');(0, '21 conv', 1, 'gru')] \\
MiddlePhalanxOutlineAgeGroup & 1 & 3 & 2 & 24 & 0.14 & 0.565 & {[}(-2, '21 dconv', -2, 'gru')] \\
MiddlePhalanxOutlineCorrect & 1 & 4 & 2 & 20 & 2.27 & 0.777 & {[}(-2, '7:4dr conv', -1, '21 conv')] \\
MiddlePhalanxTW & 1 & 3 & 2 & 24 & 0.42 & 0.519 & {[}(-2, '21 dconv', -1, '13 conv')] \\
NonInvasiveFetalECGThorax1 & 4 & 3 & 2 & 36 & 2.97 & 0.943 & {[}(-2, '21 conv', -2, '2 maxpool');(-2, '21 conv', -2, '2 maxpool');(0, '21 conv', 0, 'lstm');(0, '2 avgpool', 2, '2 maxpool')] \\
NonInvasiveFetalECGThorax2 & 4 & 3 & 4 & 24 & 1.64 & 0.945 & {[}(-2, 'identity', -2, '21 conv');(-2, '7:4dr conv', 0, '7:4dr conv');(-2, '21 dconv', 0, '7:4dr conv');(2, '21 dconv', 2, '7:4dr conv')] \\
PhalangesOutlinesCorrect & 4 & 3 & 2 & 24 & 0.26 & 0.843 & {[}(-2, '2 maxpool', -2, '2 maxpool');(-2, '7 conv', 0, '2 maxpool');(0, 'identity', 1, '2 maxpool');(-2, '7:4dr conv', 2, '13 dconv')] \\
ProximalPhalanxOutlineAgeGroup & 1 & 4 & 2 & 42 & 3.01 & 0.849 & {[}(-2, '13 dconv', -1, '7:4dr conv')] \\
ProximalPhalanxOutlineCorrect & 1 & 4 & 2 & 24 & 1.81 & 0.876 & {[}(-2, '21 conv', -2, '7:2dr conv')] \\
ProximalPhalanxTW & 2 & 4 & 2 & 24 & 1.61 & 0.776 & {[}(-2, '7 dconv', -2, '7:2dr conv');(-2, '7:2dr conv', 0, '7:2dr conv')] \\
RefrigerationDevices & 1 & 3 & 4 & 48 & 2.07 & 0.520 & {[}(-2, '13 dconv', -1, 'lstm')] \\
ScreenType & 1 & 3 & 3 & 48 & 1.03 & 0.549 & {[}(-2, '7 conv', -2, '7:4dr conv')] \\
SemgHandMovementCh2 & 2 & 3 & 2 & 24 & 0.26 & 0.771 & {[}(-2, '2 avgpool', -2, '2 avgpool');(-2, '7:2dr conv', 0, 'gru')] \\
SemgHandSubjectCh2 & 3 & 4 & 2 & 24 & 2.13 & 0.878 & {[}(-2, '7:2dr conv', -2, '2 maxpool');(-2, '2 avgpool', 0, '7:4dr conv');(-2, '7:2dr conv', 1, 'gru')] \\
ShapesAll & 4 & 3 & 4 & 20 & 2.49 & 0.897 & {[}(-2, '21 conv', -1, 'gru');(-2, '21 dconv', -1, '21 dconv');(-2, '21 conv', 1, 'gru');(-2, '13 dconv', -1, 'lstm')] \\
SmallKitchenAppliances & 1 & 3 & 2 & 24 & 0.57 & 0.803 & {[}(-2, 'gru', -1, '13 conv')] \\
StarLightCurves & 1 & 3 & 2 & 24 & 0.24 & 0.977 & {[}(-2, '21 dconv', -1, 'gru')] \\
Strawberry & 5 & 3 & 4 & 24 & 1.45 & 0.970 & {[}(-2, '7:4dr conv', -2, '7:4dr conv');(-2, '7:2dr conv', -2, 'gru');(0, '2 maxpool', 0, 'gru');(0, 'identity', 0, 'gru');(2, '2 maxpool', 3, 'gru')] \\
SwedishLeaf & 3 & 3 & 4 & 24 & 2.67 & 0.962 & {[}(-2, '13 conv', -2, '21 conv');(-2, '13 conv', -2, '21 conv');(-2, '13 conv', -2, '13 conv')] \\
TwoPatterns & 1 & 3 & 2 & 24 & 0.03 & 1.000 & {[}(-2, 'identity', -2, '7 dconv')] \\
UWaveGestureLibraryAll & 3 & 3 & 2 & 24 & 0.91 & 0.970 & {[}(-2, '7 conv', -2, 'lstm');(-2, '7:4dr conv', 0, '7:4dr conv');(-2, '13 conv', 0, '7:4dr conv')] \\
UWaveGestureLibraryX & 4 & 3 & 2 & 24 & 0.56 & 0.824 & {[}(-2, 'identity', -2, '2 maxpool');(0, '2 maxpool', 0, '2 avgpool');(-2, '7:4dr conv', 1, '7:4dr conv');(-2, '7:4dr conv', 0, 'gru')] \\
UWaveGestureLibraryY & 5 & 3 & 2 & 36 & 2.36 & 0.767 & {[}(-2, '2 maxpool', -2, '2 maxpool');(-2, '2 avgpool', 0, '7:4dr conv');(-2, '7:4dr conv', 0, '7:4dr conv');(-2, '7:4dr conv', 1, '7:4dr conv');(-2, '21 conv', 3, '2 maxpool')] \\
UWaveGestureLibraryZ & 1 & 3 & 4 & 48 & 2.91 & 0.745 & {[}(-1, '7:4dr conv', -1, 'gru')] \\
Wafer & 1 & 3 & 2 & 24 & 0.03 & 0.998 & {[}(-2, 'identity', -2, '13 dconv')] \\
\midrule
FaceDetection & 4 & 3 & 2 & 42 & 2.36 & 0.675 & {[}(-2, '7:4dr conv', -2, '2 avgpool');(-2, '7:4dr conv', 0, '2 avgpool');(-2, '7:4dr conv', 1, '7 dconv');(-2, '13 conv', 2, '7:4dr conv')] \\
LSST & 3 & 4 & 2 & 24 & 2.65 & 0.723 & {[}(-2, '7:4dr conv', -2, 'gru');(-2, '7:4dr conv', -2, 'gru');(-2, '13 dconv', -2, 'gru')] \\
PenDigits & 1 & 3 & 3 & 48 & 1.27 & 0.989 & {[}(-2, '13 dconv', -1, 'gru')] \\
PhonemeSpectra & 4 & 4 & 3 & 20 & 2.26 & 0.382 & {[}(-2, '7:4dr conv', -2, 'gru');(-2, '7 dconv', 0, 'gru');(-2, '13 dconv', -2, '7:2dr conv');(-2, 'lstm', 2, '2 maxpool')] \\
\bottomrule
\end{tabularx}
\end{table}

\begin{table}[t]
\caption{Comparison between POPNASv3 accuracy and state-of-the-art methods on the extended UCR / UEA datasets selection. The test accuracy is computed on the default test split. The rank achieved by POPNASv3 on each dataset is reported in brackets after the accuracy value.}
\label{TAB:ext-TSC-comparison}
\begin{tabularx}{\linewidth}{@{} l YYYl @{}} 
\toprule
Dataset & HC2 & InceptionTime & ROCKET & POPNASv3 \\
\midrule
Adiac & 0.813 & \textbf{0.849} & 0.785 & 0.821 (2) \\
ChlorineConcentration & 0.771 & \textbf{0.876} & 0.812 & 0.810 (3) \\
CricketX & 0.826 & \textbf{0.846} & 0.828 & 0.821 (4) \\
CricketY & 0.851 & 0.849 & 0.856 & \textbf{0.862} (1) \\
CricketZ & \textbf{0.864} & 0.854 & 0.854 & 0.831 (4) \\
Crop & 0.765 & \textbf{0.797} & 0.751 & 0.782 (2) \\
DistalPhalanxOutlineAgeGroup & \textbf{0.748} & 0.734 & \textbf{0.748} & 0.691 (4) \\
DistalPhalanxOutlineCorrect & 0.775 & \textbf{0.793} & 0.772 & \textbf{0.793} (1) \\
DistalPhalanxTW & 0.705 & 0.669 & \textbf{0.719} & 0.712 (2) \\
ECG5000 & \textbf{0.947} & 0.940 & 0.946 & 0.927 (4) \\
EOGHorizontalSignal & \textbf{0.657} & 0.633 & 0.649 & 0.655 (2) \\
EOGVerticalSignal & 0.569 & 0.511 & 0.555 & \textbf{0.591} (1) \\
ElectricDevices & \textbf{0.758} & 0.722 & 0.726 & 0.744 (2) \\
EthanolLevel & 0.696 & 0.842 & 0.578 & \textbf{0.850} (1) \\
FaceAll & 0.868 & 0.796 & 0.940 & \textbf{0.956} (1) \\
FiftyWords & 0.833 & 0.848 & 0.835 & \textbf{0.862} (1) \\
FordA & 0.954 & \textbf{0.964} & 0.946 & 0.953 (3) \\
FordB & 0.831 & \textbf{0.857} & 0.806 & 0.843 (2) \\
HandOutlines & \textbf{0.938} & \textendash & \textendash & 0.927 (2) \\
LargeKitchenAppliances & \textbf{0.907} & 0.899 & 0.893 & 0.904 (2) \\
MedicalImages & \textbf{0.808} & 0.795 & 0.799 & 0.784 (4) \\
MiddlePhalanxOutlineAgeGroup & \textbf{0.597} & 0.558 & 0.584 & 0.565 (3) \\
MiddlePhalanxOutlineCorrect & \textbf{0.852} & 0.828 & 0.828 & 0.777 (4) \\
MiddlePhalanxTW & \textbf{0.558} & 0.506 & \textbf{0.558} & 0.519 (3) \\
NonInvasiveFetalECGThorax1 & \textbf{0.947} & \textendash & \textendash & 0.943 (2) \\
NonInvasiveFetalECGThorax2 & \textbf{0.967} & \textendash & \textendash & 0.945 (2) \\
PhalangesOutlinesCorrect & 0.831 & \textbf{0.857} & 0.831 & 0.843 (2) \\
ProximalPhalanxOutlineAgeGroup & \textbf{0.854} & 0.849 & \textbf{0.854} & 0.849 (3) \\
ProximalPhalanxOutlineCorrect & 0.900 & \textbf{0.931} & 0.900 & 0.876 (4) \\
ProximalPhalanxTW & \textbf{0.829} & 0.771 & 0.805 & 0.776 (3) \\
RefrigerationDevices & 0.539 & 0.528 & \textbf{0.544} & 0.520 (4) \\
ScreenType & 0.579 & \textbf{0.581} & 0.499 & 0.549 (3) \\
SemgHandMovementCh2 & \textbf{0.856} & 0.569 & 0.653 & 0.771 (2) \\
SemgHandSubjectCh2 & \textbf{0.902} & 0.764 & 0.869 & 0.878 (2) \\
ShapesAll & 0.922 & \textbf{0.928} & 0.910 & 0.897 (4) \\
SmallKitchenAppliances & \textbf{0.837} & 0.773 & 0.813 & 0.803 (3) \\
StarLightCurves & \textbf{0.982} & 0.978 & 0.980 & 0.977 (4) \\
Strawberry & 0.976 & \textbf{0.984} & 0.981 & 0.970 (4) \\
SwedishLeaf & 0.965 & \textbf{0.971} & 0.970 & 0.962 (4) \\
TwoPatterns & \textbf{1.000} & \textbf{1.000} & \textbf{1.000} & \textbf{1.000} (1) \\
UWaveGestureLibraryAll & 0.974 & 0.950 & \textbf{0.975} & 0.970 (3) \\
UWaveGestureLibraryX & \textbf{0.855} & 0.824 & 0.854 & 0.824 (3) \\
UWaveGestureLibraryY & \textbf{0.775} & 0.767 & 0.772 & 0.767 (3) \\
UWaveGestureLibraryZ & \textbf{0.797} & 0.768 & 0.796 & 0.745 (4) \\
Wafer & \textbf{1.000} & 0.999 & 0.998 & 0.998 (3) \\
\midrule
FaceDetection & 0.660 & \textendash & 0.644 & \textbf{0.675} (1) \\
LSST & 0.643 & 0.612 & 0.637 & \textbf{0.723} (1) \\
PenDigits & 0.979 & 0.988 & 0.983 & \textbf{0.989} (1) \\
PhonemeSpectra & 0.290 & \textendash & 0.276 & \textbf{0.382} (1) \\
\bottomrule
\end{tabularx}
\end{table}

\clearpage

\section{Appendix}
\subsection{Predictor performance on different hardware} \label{SEC:Apx-B-hw-predictors}

POPNASv3 algorithm supports different hardware configurations for training the candidate neural network models.
The underlying training procedure is carried out using the distribution strategies implemented in the TensorFlow API, enabling our algorithm to train the architectures on a single GPU, multiple GPUs, or even TPU devices.
The training time of the models can vary in a non-linear manner with the available computational power of the hardware, so we perform some experiments to study the robustness of our time predictor and feature set to different hardware environments.
The predictors are fundamental for guiding the algorithms into exploring the most promising regions of the search space, as the Pareto front depends on the correctness of the ranking of the evaluated architectures.
Therefore, the predictors must be able to rank correctly the results not only for different tasks but also for diverse training techniques.

We perform three experiments on the CIFAR10 dataset, using a different hardware configuration for the network training and adapting the learning rate and batch size based on the device.
The first experiment is the one described in Section~\ref{SEC:IC-results}, performed on a single A100 MIG partition (3g.40gb profile) using the default hyperparameters listed in Table~\ref{TAB:hyperparameters}.
The other two experiments are performed respectively on 2 A100 GPUs and an 8 cores TPUv2 machine, increasing in both experiments the batch size to 256, to exploit better the available devices.
We observed in a previous ablation study on the TPU that increasing the batch size to 512 or more leads to generally worse accuracy in all networks, so we limit it to 256 even if TPUs can exploit much higher parallelization.
The starting learning rate is adapted to 0.02, following the heuristic of multiplying the learning rate for the same factor of the batch size.

\begin{figure}[t]
	\centering
	\subfloat[A100 (MIG 3g.40gb)]{
	    \label{FIG:a100-acc}
	    \includegraphics[width=0.32\textwidth]{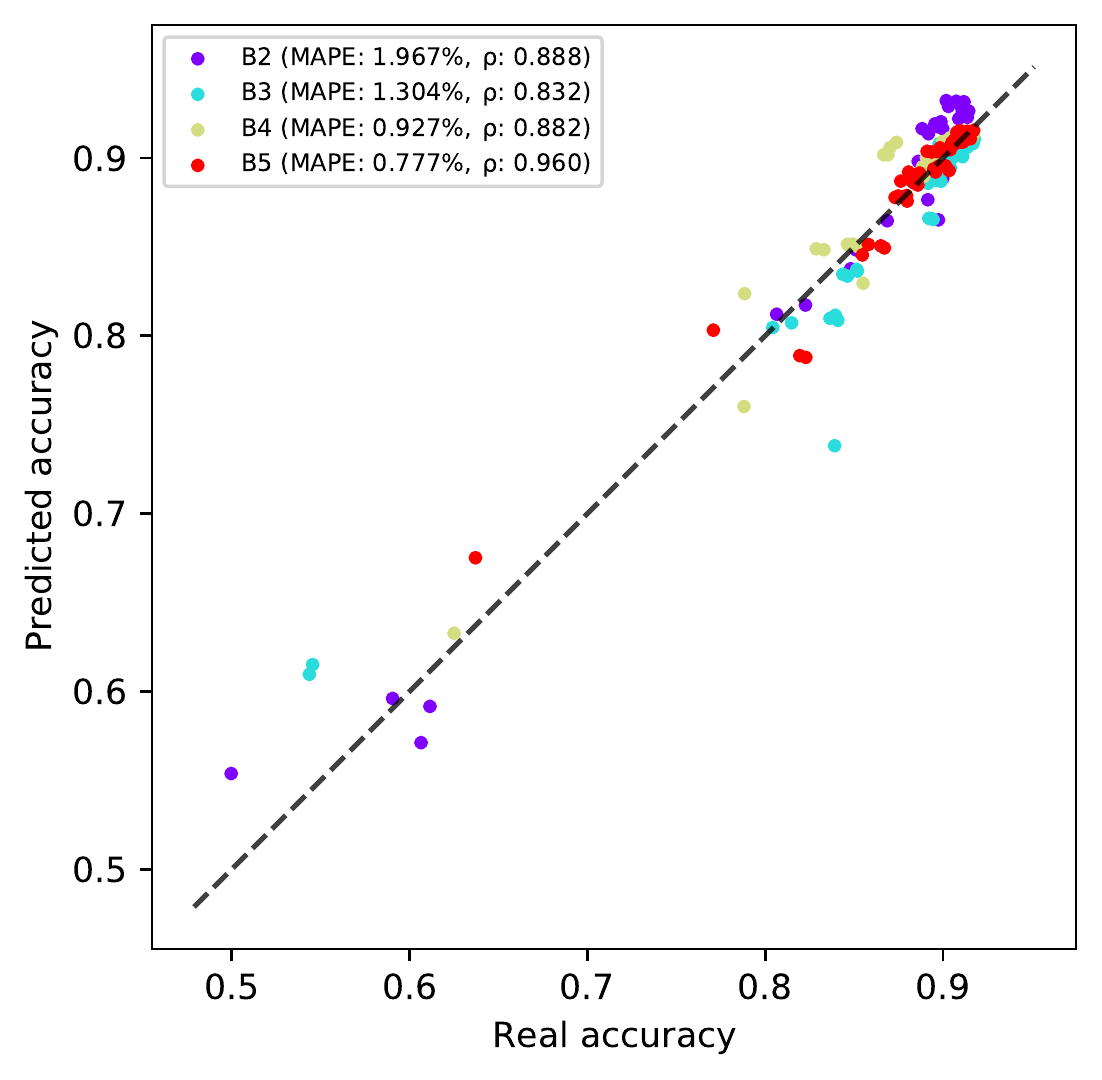}
	}
	\hfill
	\subfloat[2xA100]{
	    \label{FIG:2x-a100-acc}
	    \includegraphics[width=0.32\textwidth]{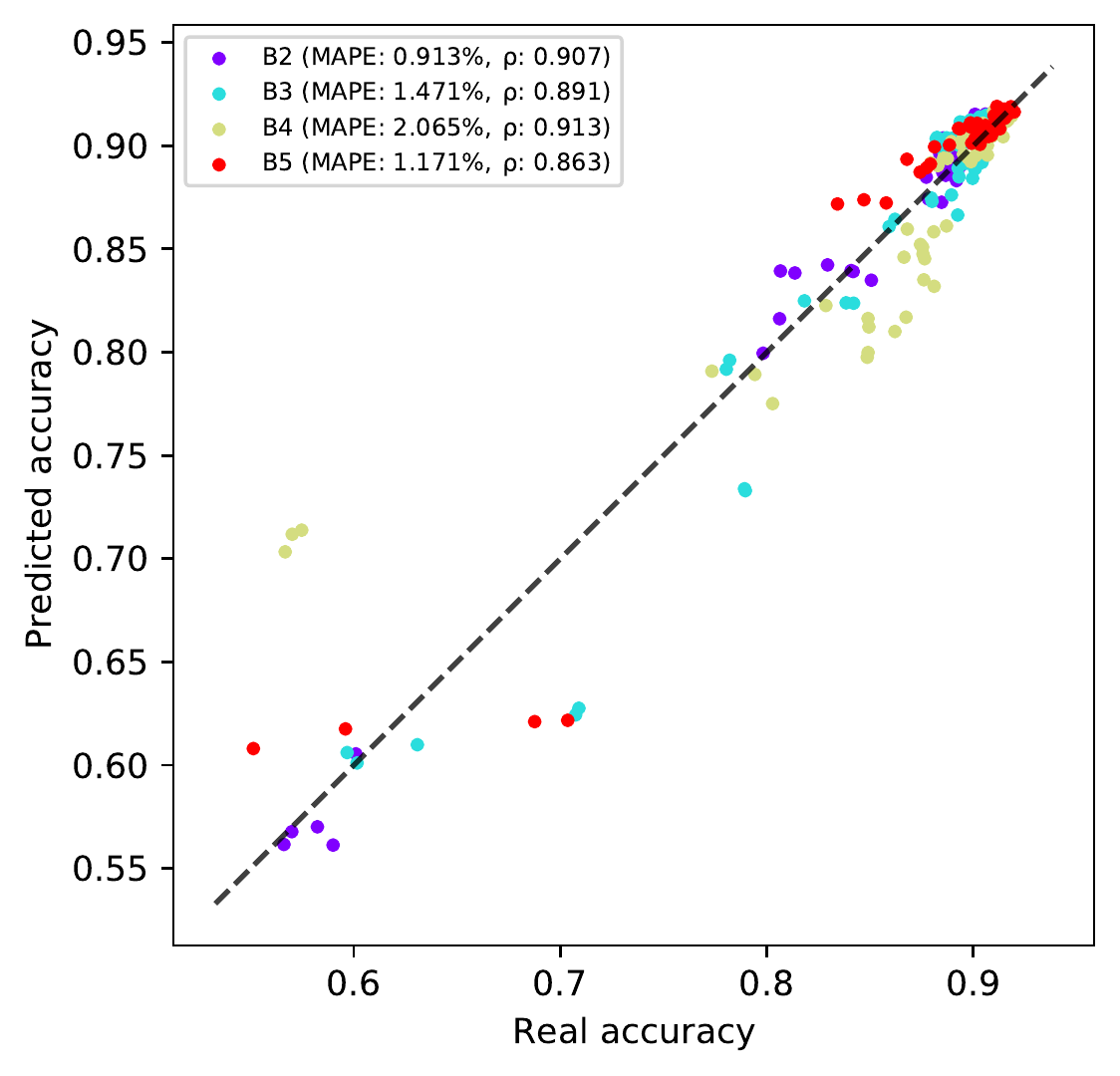}
	}
    \hfill
	\subfloat[TPU]{
	    \label{FIG:tpu-acc}
	    \includegraphics[width=0.32\textwidth]{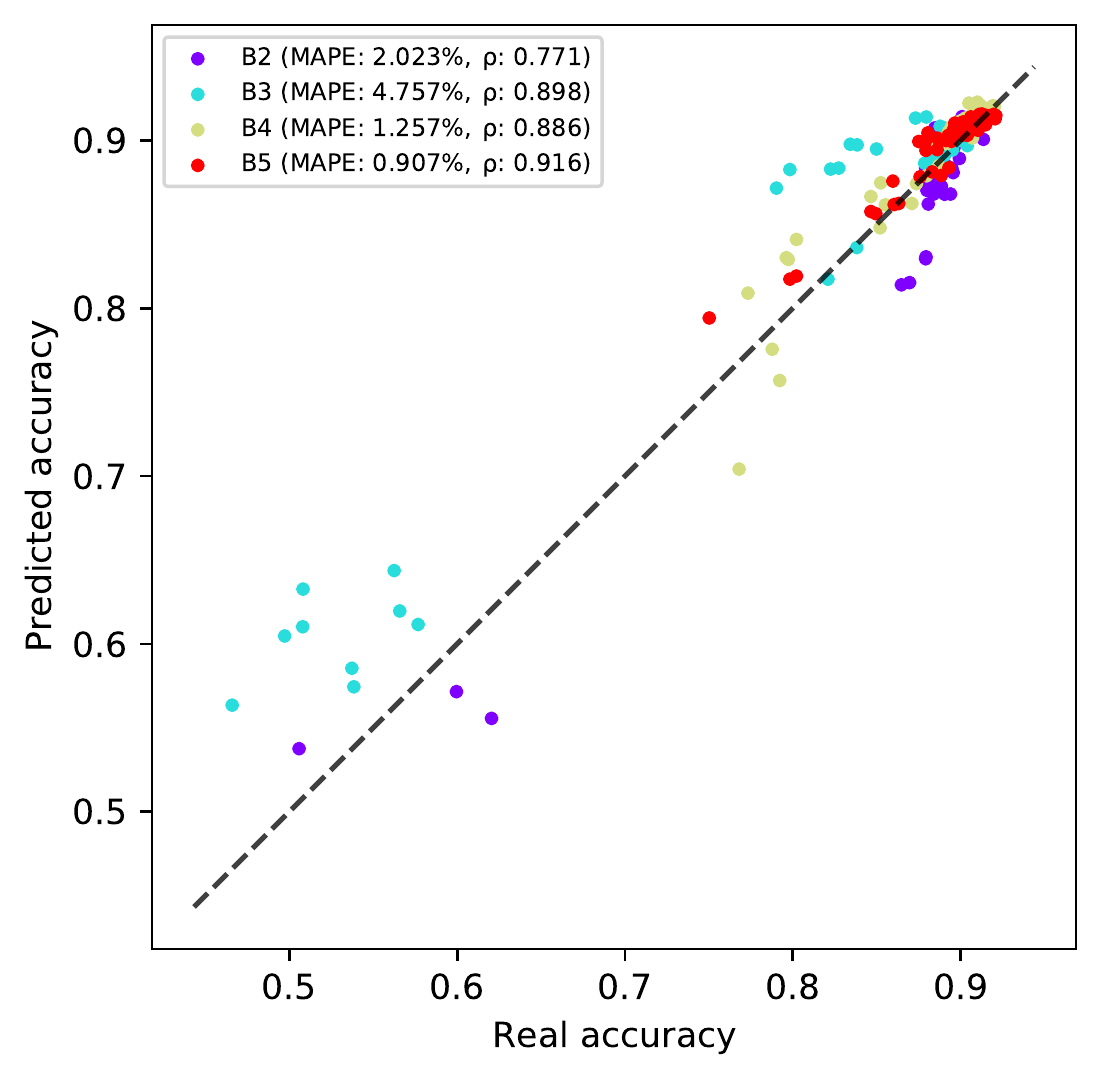}
	}
	\caption{Scatter plot representing the predicted accuracy compared to the real one retrieved after training, for each hardware-related experiment.}
	\label{FIG:acc-predictors-hw-comparison}
\end{figure}

\begin{figure}[t]
	\centering
	\subfloat[A100 (MIG 3g.40gb)]{
	    \label{FIG:a100-time}
	    \includegraphics[width=0.32\textwidth]{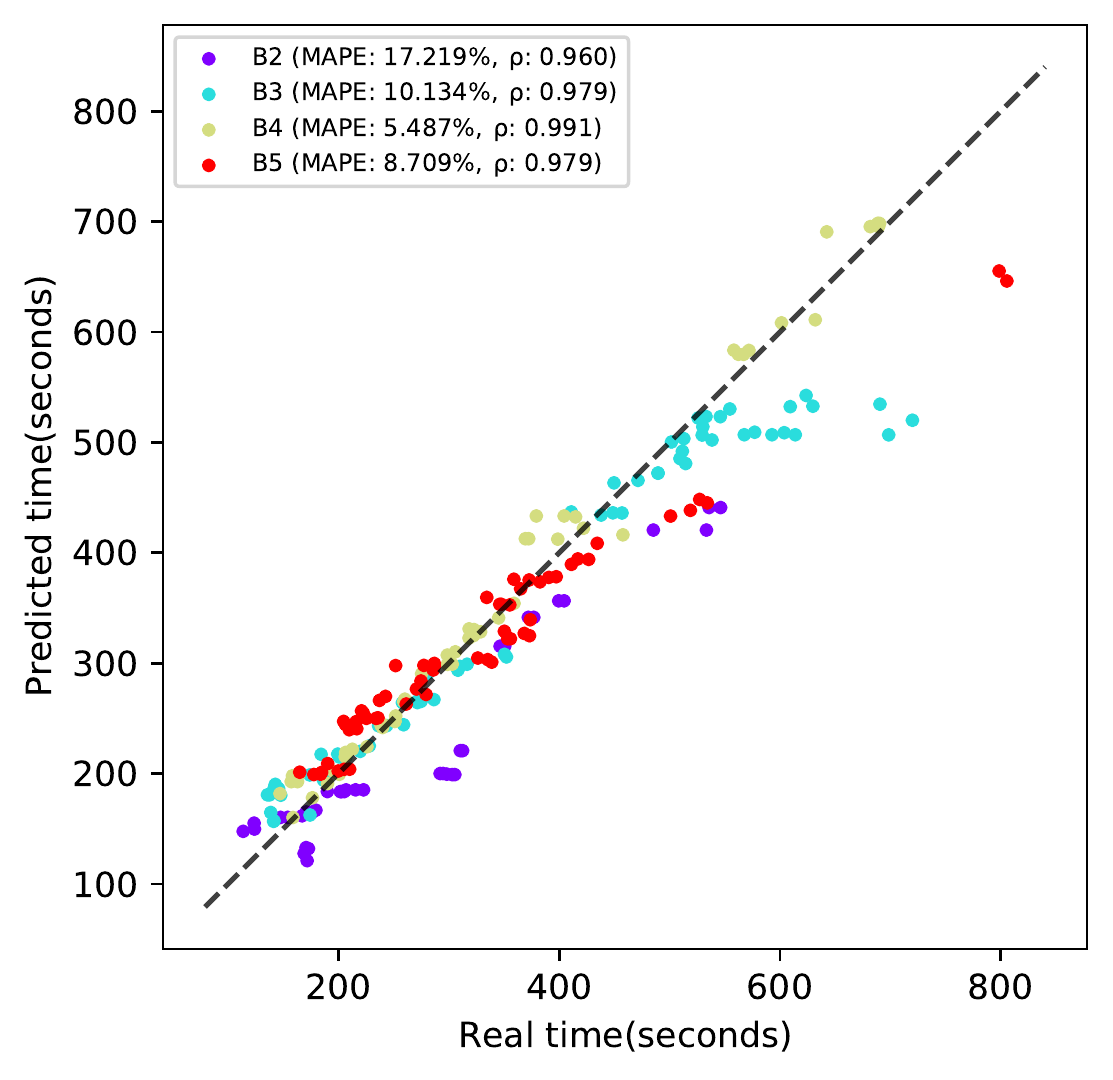}
	}
	\hfill
	\subfloat[2xA100]{
	    \label{FIG:2x-a100-time}
	    \includegraphics[width=0.32\textwidth]{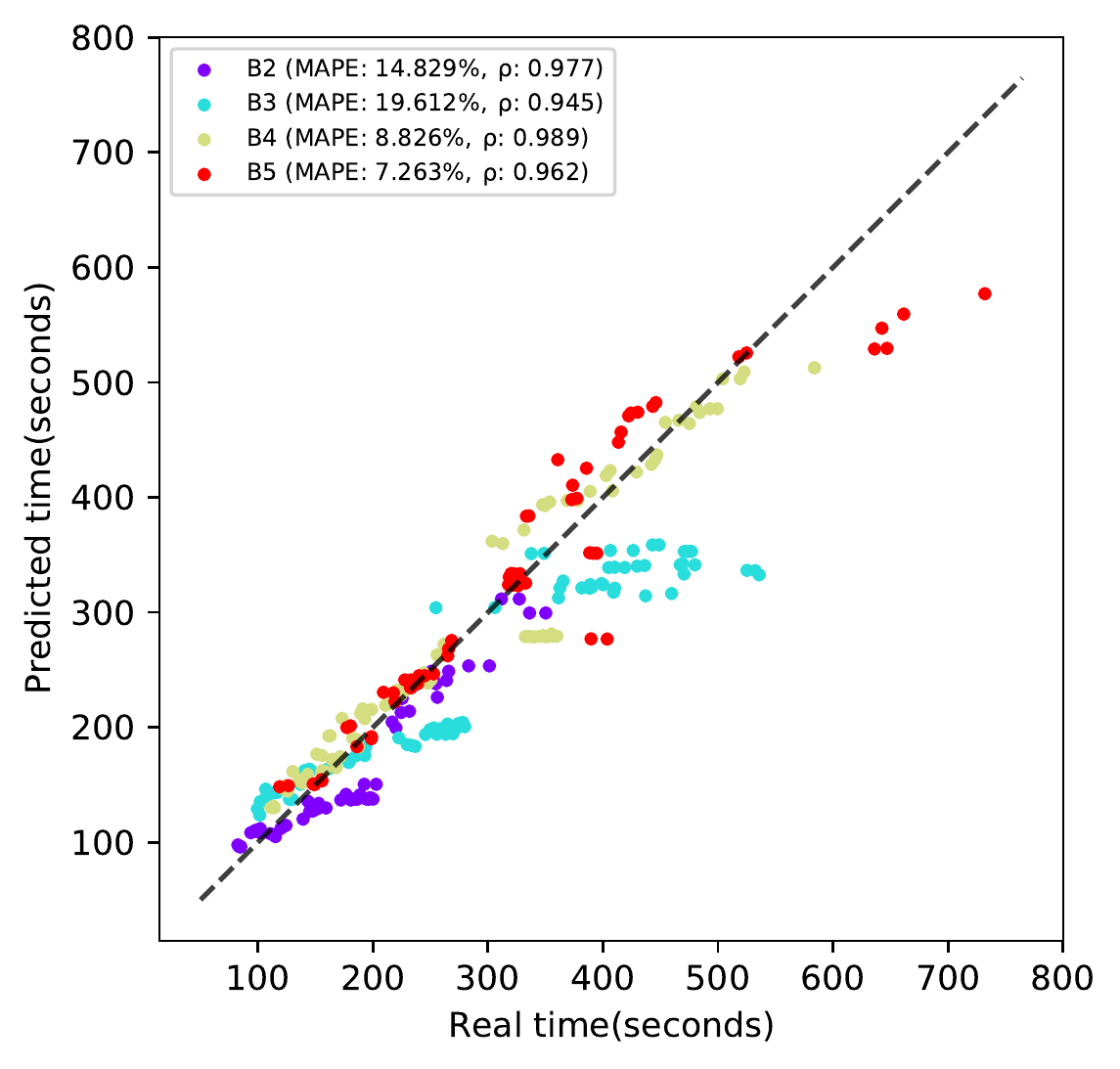}
	}
    \hfill
	\subfloat[TPU]{
	    \label{FIG:tpu-time}
	    \includegraphics[width=0.32\textwidth]{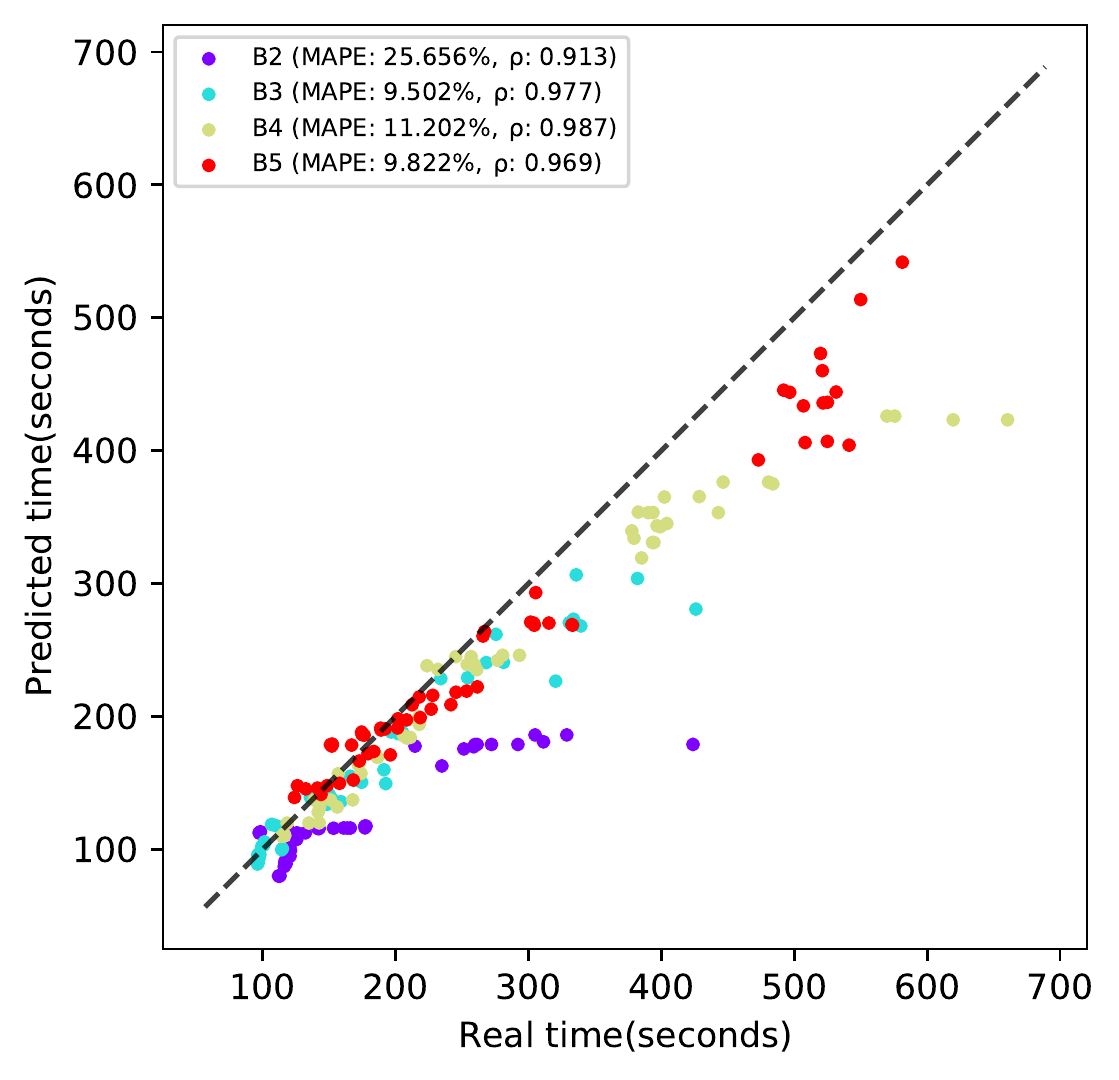}
	}
	\caption{Scatter plot representing the predicted training time compared to the real one retrieved after training, for each hardware-related experiment.}
	\label{FIG:time-predictors-hw-comparison}
\end{figure}

The results of the predictors on these experiments are presented in Table~\ref{TAB:hw_acc_predictors_comparison} and Table~\ref{TAB:hw_time_predictors_comparison}.
A visual comparison is also available in the scatter plots presented in Figure~\ref{FIG:acc-predictors-hw-comparison} and Figure~\ref{FIG:time-predictors-hw-comparison}.
The accuracy reached by the networks should not depend on the hardware used for training, but it can be slightly affected by the batch size and learning rate changes.
The accuracy predictor has similar performance in the three experiments due to the low variance of the accuracy reached by the architectures, which is more imputable to the weight initialization and non-deterministic operations executed in parallel architectures than the hardware configuration themselves.
The MAPE per step is minor and the Spearman's rank correlation coefficient is significantly high in all configurations, giving indications that the accuracy predictor is suited for the task.

\begin{table}[t]
	\caption{The comparison of POPNASv3 accuracy predictor results on CIFAR10, using different hardware environments during neural networks training.}
	\centering
	\begin{tabularx}{\linewidth}{@{} p{4.5cm} XXXX @{\hskip 1cm} XXXl @{}}
		\toprule
		\multirow{2}{*}{Device} & \multicolumn{4}{c}{MAPE(\%)} & \multicolumn{4}{c}{Spearman($ \rho $)} \\
		\cmidrule{2-5} \cmidrule{6-9}
		& b=2 & b=3 & b=4 & b=5 & b=2 & b=3 & b=4 & b=5 \\
		\midrule
		A100 80GB (MIG 3g.40gb) & 1.967 & 1.304 & 0.927 & 0.777 & 0.888 & 0.832 & 0.882 & 0.960 \\
        2xA100 80GB & 0.913 & 1.471 & 2.065 & 1.171 & 0.907 & 0.891 & 0.913 & 0.863 \\
        TPU v2-8 & 2.023 & 4.757 & 1.257 & 0.907 & 0.771 & 0.898 & 0.886 & 0.916 \\
		\bottomrule
	\end{tabularx}
	\label{TAB:hw_acc_predictors_comparison}
\end{table}

\begin{table}[t]
	\caption{The comparison of POPNASv3 time predictor results on CIFAR10, using different hardware environments during neural networks training.}
	\centering
	\begin{tabularx}{\linewidth}{@{} p{4.5cm} XXXX @{\hskip 1cm} XXXl @{}}
		\toprule
		\multirow{2}{*}{Device} & \multicolumn{4}{c}{MAPE(\%)} & \multicolumn{4}{c}{Spearman($ \rho $)} \\
		\cmidrule{2-5} \cmidrule{6-9}
		& b=2 & b=3 & b=4 & b=5 & b=2 & b=3 & b=4 & b=5 \\
		\midrule
		A100 80GB (MIG 3g.40gb) & 17.219 & 10.134 & 5.487 & 8.709 & 0.960 & 0.979 & 0.991 & 0.979 \\
        2xA100 80GB & 14.829 & 19.612 & 8.826 & 7.263 & 0.977 & 0.945 & 0.989 & 0.962 \\
        TPU v2-8 & 25.656 & 9.502 & 11.202 & 9.822 & 0.913 & 0.977 & 0.987 & 0.969 \\
		\bottomrule
	\end{tabularx}
	\label{TAB:hw_time_predictors_comparison}
\end{table}

The time predictor results are instead more relevant, since the training time speed-up could be different for each architecture and therefore the predictor must adapt to the hardware environment.
Analyzing the architectures with cells composed of a single block, which are fixed by our training procedure and therefore the same in all experiments, we notice a $(1.47\pm0.09)$ average speed-up between using a partitioned A100 and two A100 with replicated strategy, while using TPU increases this factor to a $(1.47\pm0.26)$ speed-up.
The speed-up standard deviation is not negligible in both cases, which means that the time predictor must adapt to a non-linear speed-up.
The dynamic reindex formula is tuned on the results obtained at runtime, making it possible for the predictor to learn training time discrepancies between the operators in different hardware environments.
The results indicate that the time predictor preserves a Spearman rank coefficient close to 1 in every step, for all experiments.
The ranking of the training time is therefore optimal also on different training devices, demonstrating that our method can work well for heterogeneous hardware and training strategies.

\end{document}